\documentclass[10pt]{article}
\usepackage[utf8]{inputenc}
\usepackage{amsmath, amssymb}
\usepackage{graphicx}
\usepackage{float}
\usepackage{algorithmicx} 
\usepackage{caption}
\usepackage{geometry}
\usepackage{multicol}
\usepackage{fancyhdr}
\usepackage[numbers]{natbib}
\usepackage[ruled,lined,linesnumbered]{algorithm2e}
\SetKwComment{Comment}{/* }{ */}
\SetKwInput{KwSteps}{Steps}
\SetKwFor{for}{for}{do}{end}
\SetNlSty{texttt}{}{.}
\newtheorem{proposition}{Proposition}
\newcommand{\proof}{{\it Proof.} }
\makeatletter
\renewcommand{\abstractname}{\centering \normalsize Abstract}
\renewenvironment{abstract}{
    \begin{center}
    \bfseries \abstractname\vspace{-.5em}\vspace{0pt}
    \end{center}
    \quotation
}{\endquotation}
\makeatother

\newenvironment{keywords}{
  \begin{center}
  \bfseries \normalsize Keywords
  \end{center}
  \quotation
}{\endquotation}

\title{\bfseries Modification of a Numerical Method Using FIR Filters in a Time-dependent SIR Model for COVID-19}
\author{Felipe Rog\'erio Pimentel \\ \small fpimentel@ufop.edu.br \\ 
\small \textit{Department of Mathematics,} \\ \small \textit{Federal University of Ouro Preto, Ouro Preto, Brazil.}
\and Rafael Gustavo Alves \\ \small rafael.gustavo.alves@gmail.com }

\date{} 

\begin{document}

\twocolumn[
    \maketitle
    \begin{abstract}
        Authors Yi-Cheng Chen, Ping-En Lu, Cheng-Shang Chang, and Tzu-Hsuan Liu use the Finite Impulse Response (FIR) linear system filtering method to track and predict the number of people infected and recovered from COVID-19, in a pandemic context in which there was still no vaccine and the only way to avoid contagion was isolation. To estimate the coefficients of these FIR filters, Chen et al. used machine learning methods through a classical optimization problem with regularization (ridge regression). These estimated coefficients are called ridge coefficients. The epidemic mathematical model adopted by these researchers to formulate the FIR filters is the time-dependent discrete SIR. In this paper, we propose a small modification to the algorithm of Chen et al. to obtain the ridge coefficients. We then used this modified algorithm to track and predict the number of people infected and recovered from COVID-19 in the state of Minas Gerais/Brazil, within a prediction window, during the initial period of the pandemic. We also compare the predicted data with the respective real data to check how good the approximation is. In the modified algorithm, we set values for the FIR filter orders and for the regularization parameters, both different from the respective values defined by Chen et al. in their algorithm. In this context, the numerical results obtained by the modified algorithm in some simulations present better approximation errors compared to the respective approximation errors presented by the algorithm of Chen et al.
    \end{abstract}
    \vspace{1em}
]
\normalsize
\section{Introdução}
\label{introd}
Mathematical epidemiological models have been widely used to describe the dynamics of infectious-disease transmission. There are two types of epidemic model in the literature: continuous and discrete. Continuous-time models use a theoretical mathematical approach, whereas discrete-time models use a preferable numerical approach.
Among the classical models studied, the most notable and important is the classical Kermack-McKendric SIR model \cite{KM, KM2}, from which most other infectious disease models have emerged. Generally speaking, the susceptible class (S) consists of people who may have the disease but have not yet been infected. The infected class (I) consists of those who are the transmitters of the disease to other people. The removed class (R) from the S-I interaction consists of people who have recovered from the disease with immunity, isolation, or death. 

Some infectious disease models have specific compartments for the class of people with temporary immunity (M) and for the class of people exposed (E) to the disease, that is, people in a latent period in which they can infect others without having disease symptoms. Some of these models are of the MSEIR type, have been extensively studied by Hethcote \cite{HH2}, and can be applied to diseases such as AIDS, COVID-19, measles, rubella, and mumps. 

Other models studied include SI, SIS, SIRS, SEIR, SEIRS, MSEIRS, SEI, and SEIS \cite{KM, KM2, HH, HH2, BT}. Anderson and May \cite{AM} discussed an SII-type model for directly transmitted microparasitic infections, the compartment model of which is described by Susceptible, Infected and Immune hosts. The discussion is extended in the second part of the article \cite{AM2} to microparasites such as viruses, bacteria, and protozoa, and to macroparasites such as helminths and arthropods, in which disease transmission occurs directly or indirectly, through one or more intermediate hosts.

Many models assume that the total population size is constant \cite{HH, HH2, ALLEN, AJM, CHEN, YWXL} and/or take into account vital dynamics \cite{HH, ALLEN, BT}; that is, they consider births and deaths of people. However, in order to consider that the total population remains constant over a given period of time of study of disease transmission, it is necessary to assume that the birth rate is equal to the rate of natural deaths (not due to disease). Models dealing with deaths caused by disease are also studied, such as the SIS model with some disease fatalities studied by Hethcote \cite{HH} and the SIRD model studied by Teran \cite{CT}, but the rate of these deaths is treated separately from the rate of natural deaths. As examples of models with vital dynamics, we can mention those studied by Hethcote \cite{HH}: SIR with carriers (hepatitis, polio, diptheria, typhoid fever, and cholera), SIRS with temporary immunity (smallpox, tetanus, influenza, cholera, and typhoid fever), and SIS with migration (in which individuals are assumed to immigrate and emigrate between two communities at equal rates). The vital dynamics SIS model was also studied by Allen \cite{ALLEN}, in the discrete-time SIS version.

However, there are also models that consider the total population varying over time. Hethcote \cite{HH2} formulates a system of differential equations for the MSEIR model in which he assumes a population with exponentially varying size. Beretta and Takeuchi \cite{BT} attempt to improve the convergence results obtained in their previous paper \cite{BT2} in a generalized SIR model with variable population size, vital dynamics and distributed incubation times. Allen \cite{ALLEN} formulated a discrete-time SIS model with variable population size. 

Models that take vaccination into account should also be mentioned. These models adopt a specific compartment for the class of vaccinated individuals (V). Hubert and Turinici \cite{HT} study a SIRV model with vital dynamics, in a context of non-compulsory vaccination of newborns, and in which the evolution of each individual is modeled as a Markov chain. Allen \textit{et al.} \cite{AJM} analyze a discrete-time SIR model that accounts for the effects of vaccination and that is applied to a localized measles epidemic on a university campus. In this model, they observe three important mathematical properties: the non-negativity of the solutions, the invariance of the total population size over time, and the end of the epidemic occurring with all individuals remaining susceptible or becoming immune. There are also SIRVS-type epidemic models that assume that the class of recovered people has permanent immunity and the class of vaccinated people has temporary immunity, which means that vaccinated people can lose immunity and become susceptible again. Among the infectious diseases described by this type of model we can mention: measles, rubella, chickenpox, hepatitis A, etc. COVID-19 can also be described using a SIRVS-type model, although it is known that those who have recovered from the disease are not permanently immune. Zhang  \textit{et al.} \cite{ZTG} study the continuous version of the SIRVS model, while Zhang \cite{ZHANG} studies its discrete version, both studies under the assumption that the coefficients (or parameters) are also time dependent. 

There is another class of models that takes into account the time delay, represented by a time parameter that indicates the incubation period of the disease, i.e., the time between infection and symptom onset \cite{WZJ, BT, CT, RAR}. 

Finally, we mention some models applied to COVID-19. Ketcheson \cite{KET} models the spread of COVID-19 through the classic SIR model in which the system equations also include a variable that describes the fraction of people who adopt positions such as isolation and social distancing with the aim of reducing the contagion rate and thus decreasing the spread of the disease. Chen  \textit{et al.} \cite{CHEN} propose a time-dependent SIR model that tracks the transmission and recovery rate in discrete time to predict the trend of COVID-19 using data provided by the China authority. Teran \cite{CT} develops a discrete-time SIRD model with time delay for the spread of COVID-19 to predict the number of infected, recovered and deceased persons in the Peruvian regions of Moquegua and Tacna. His model is based on the Chen  \textit{et al.} \cite{CHEN} model and uses functional differential equations. Ivorra \textit{et al.} \cite{IVR} developed a model called $\theta-\textrm{SEIHRD}$ that takes into account the known special characteristics of COVID-19 in the most critical years of the disease, such as the existence of undetected infectious cases and the different health and infectiousness conditions of hospitalized people. The studies were aimed at the particular case of China. Rihan  \textit{et al.} 
 \cite{RAR} use a stochastic epidemic SIRC model with cross-immunity class and time delay to study the spread of COVID-19. They analyze the model and prove the existence and uniqueness of a positive global solution. The SIRC model was introduced by Casagrandi  \textit{et al.} \cite{CBLA} by inserting the compartment C called cross-immunity to describe the dynamics of Influenza A among people who, according to Rihan  \textit{et al.} \cite{RAR},``recovered after being infected by different strains of the same viral subtype in previous years. The C compartment describes an intermediate state between Susceptible and Recovered.'' Wacker and Schl\"uter \cite{WS} revisit continuous-time and discrete-time SIR models and use their proposed discrete-time SIR model to study the spread of COVID-19 in Germany and Iran. All these applied epidemic models were developed in 2019-2020, the most critical years of COVID-19. 

Next, we present an overview of this paper. In Section \ref{sirmodel} we formulate the time-dependent SIR model in both continuous-time and discrete-time versions. Although it is a model that can be applied to other infectious diseases, we will focus specifically on COVID-19, in the early period of the disease in which the only measures available to prevent transmission were isolation and social distancing. In Section \ref{equivalence} we show the equivalence between the discrete models in both situations: one in which we treat the class of people who died from the disease and the class of recovered people in the same compartment, and the other in which we treat these classes in separate compartments. In Section \ref{estimating} we formulate, as in \cite{CHEN}, two optimization problems with regularization whose solutions (ridge coefficients) are used to obtain estimates of the transmission and recovery rates, respectively, within a prediction window. We also demonstrate the proposition that establishes the solution of these two problems whose proof uses Singular Value Decomposition. In Section \ref{modification} we show our algorithm which is a modification of a version of the algorithm of Chen  \textit{et al.} \cite{CHEN} and present the results of our numerical simulations with data obtained by the health department of the state of Minas Gerais/Brazil. The conclusion of the paper is presented in Section \ref{conclusion}.
\section{The Time-dependent SIR model}
\label{sirmodel}

We begin this section from the differential equations for the continuous-time version of the Time-dependent SIR model addressed by Chen  \textit{et al.} \cite{CHEN} to model the spread of COVID-19 in a context of lack of vaccination:
\begin{align}\label{sistTDSIR}
	S' &= -\dfrac{\beta(t) S(t)I(t)}{n}, \nonumber \\
	I' &= \dfrac{\beta(t) S(t)I(t)}{n} - \gamma(t) I(t), \\
	R' &= \gamma(t) I(t).\nonumber
\end{align}
where
\begin{equation}\label{const} S(t) + I(t) + R(t) \approx n. 
\end{equation} 
Here $n$ represents the total population which we suppose it is constant. The variables $S(t)$ and $I(t)$ describe the number of susceptible and infected people at time $t$. The variable $R(t)$ defines the number of recovered and dead persons from the disease at time $t$. The parameters $\beta$ and $\gamma$ describe the transmission rate and the recovery rate, respectively. The transmission rate $\beta$ is the average number of contacts per infected individual per day. The recovery rate $\gamma$ describes the daily average number of individuals removed from the infected class and who recovered or died (the reason why the number of deaths is considered together with the number of recovered persons will be explained later). Unlike the classical model SIR, the parameters $\beta$ and $\gamma$ in the time-dependent SIR model are also considered a function of time $t.$ According to Chen  \textit{et al.} \cite{CHEN} this assumption ``is much better to track the disease spread, control, and predict the future trend''.

Taking into account the daily update of data on COVID-19 during the pandemic period together with the proposal to obtain a forecast of the number of people who is infected by the disease throughout the contagion period, it is important that we address the differential equations (\ref{sistTDSIR}) in their discrete-time version whose difference equations are given by
\begin{align}\label{discretesystem}
	S(t+1) - S(t) = &\, - \dfrac{\beta(t)S(t)I(t)}{n}, \nonumber \\
	I(t+1) - I(t) = &\, \dfrac{\beta(t)S(t)I(t)}{n} - \gamma(t) I(t),  \\
	R(t+1) - R(t) = &\, \gamma(t) I(t). \nonumber
\end{align}
where the time step is $\Delta t = 1$. Here, the variable $t$ is assumed to be a discrete integer. We note that $S(t), I(t)$ and $R(t)$ still satisfy (\ref{const}).

Unlike \cite{CHEN}, we will not assume $S(t)\approx n$ although we know that most of the population is in the susceptible class at the beginning of the disease spread. We will also not work directly with the variables $S(t), I(t)$ and $R(t)$ but with their respective fractions whose denominators are the constant total population $n$ \cite{HH2}, that is  
\begin{equation}\label{fractions}
s(t) = S(t)/n, ~~ i(t) = I(t)/n, ~~ r(t) = R(t)/n.	
\end{equation}
Thus, dividing the equations in (\ref{discretesystem}) and the equation (\ref{const}) by $n,$ and using (\ref{fractions}) implies
\begin{align}
	s(t+1) - s(t) =& \, - \beta(t)s(t)i(t), \label{discrete_s} \\
	i(t+1) - i(t) =& \, \beta(t)s(t)i(t) - \gamma(t) i(t), \label{discrete_i} \\
	r(t+1) - r(t) =&\,  \gamma(t) i(t). \label{discrete_r}
\end{align}
and   
\begin{equation}\label{const2} s(t) + i(t) + r(t) \approx 1. 
\end{equation}  
We can easily express the parameters $\gamma$ and $\beta$ in terms of the variables $i$ and $r$ as follows: from (\ref{discrete_r}), we have
\begin{equation}
	\gamma(t) = \dfrac{r(t+1)- r(t)}{i(t)} \label{gammat}
\end{equation}
and, using (\ref{gammat}) and (\ref{const2}) in (\ref{discrete_i}) produces 
\begin{equation}
	\beta(t) = \dfrac{[i(t+1)- i(t)]+[r(t+1)- r(t)]}{i(t)[1-i(t)-r(t)]}. \label{betat}
\end{equation}
According to \cite{CHEN} we want to obtain predictions of the numbers of infected and recovered people for the days $t = T, T+1,...,T+W$ assuming that we know such data for the period $0\leq t\leq T-1$ . Here $W$ represents what we could call a \textit{prediction window}. Thus, from collected data $\left\{ i(t), r(t) \mid 0\leq t \leq T-1 \right\}$, we first obtain the corresponding daily rates $\left\{ \beta(t), \gamma(t) \mid 0 \leq t \leq T-2 \right\}$ by using (\ref{gammat}) and (\ref{betat}), then we can use specific algorithms built to predict $\left\{ \beta(t), \gamma(t) \mid  t \geq T-1 \right\},$ and hence 
\begin{equation}\label{set_i_r}
	\left\{ i(t), r(t) \mid  t \geq T \right\}.
\end{equation} 

To obtain the set of data in (\ref{set_i_r}) we must use (\ref{discrete_i}), (\ref{discrete_r}) and (\ref{const2}) in such a way that
\begin{align}
	i(t+1) =&\, [1+\beta(t)(1-i(t)-r(t)) - \gamma(t)] i(t), \label{i_predicted} \\
	r(t+1) =&\, r(t)+\gamma(t) i(t). \label{r_predicted}
\end{align} 

So, according to (\ref{fractions}) we obtain the values of $$\left\{ I(t), R(t) \mid  t \geq T \right\}$$ by multiplying $i(t)$ and $r(t)$ by $n.$

Before giving the details of the procedures in the previous paragraph, we will explain why we consider together the rates of those recovered and those who died from the disease. 

\section{The equivalence between models}
\label{equivalence}

Chen \textit{et al.} \cite{CHEN} say
\begin{quote}
  ``\dots The reason for the number of deaths is counted in the recovery state is that, from an epidemiological point of view, this is basically the same thing, regardless of whether recovery or death does not have much impact on the spread of the disease''.  
\end{quote}
In addition, we can give another explanation (under the assumption $s(t) \not\equiv 1, \, s $ as in (\ref{fractions})), this time from an algebraic point of view. In fact, if we consider the number of persons who died from the disease as elements of an exclusive compartment and represent this number by the variable $D(t)$ (respectively $d(t)= D(t)/n$), the system of difference equations that describes the dynamics of the disease becomes
\begin{align}
		s(t+1)-s(t)=&\, -\beta(t) s(t)i(t), \label{discrete_s2} \\
		i(t+1)-i(t)=&\,\beta(t) s(t)i(t)-\gamma(t)i(t)-\delta(t)i(t), \label{discrete_i2} \\
		r(t+1)-r(t)=&\,\gamma(t)i(t), \label{discrete_r2} \\
		d(t+1)-d(t)=&\,\delta(t)i(t), \label{discrete_d2}
\end{align}
where we are now considering  that the recovered rate $\gamma(t)$ is separate from the disease death rate $\delta(t)$ at time $t$. 

It is easy to verify that the variables $S(t), I(t), R(t)$ and $D(t)$ also satisfy $S(t)+I(t)+R(t)+D(t) \approx n \ \ \forall t,$ and, respectively. 
\begin{equation} \label{const3} s(t)+i(t)+r(t)+d(t) \approx 1 \ \ \forall t.  
\end{equation} 

From (\ref{discrete_r2}) we still have (\ref{gammat}), and from (\ref{discrete_d2}) we have
\begin{equation}
	\delta(t) = \dfrac{d(t+1)- d(t)}{i(t)}. \label{deltat}
\end{equation}
Using (\ref{gammat}), (\ref{deltat}) and (\ref{const3}) in (\ref{discrete_i2}) implies
\begin{equation}\label{betat2}
	\beta(t)=\frac{i(t+1)-i(t)+r(t+1)-r(t)+d(t+1)-d(t)}{i(t)[1-(i+r+d)(t)]}
\end{equation}

Let us now define the two sums $r+d$ and $\gamma + \delta$ by the new variables $\tilde{r}$ and $\tilde{\gamma},$ respectively. Thus, adding the equations (\ref{discrete_r2}) and (\ref{discrete_d2}) side by side and replacing $(r+d)(t)$ and $(\gamma + \delta)(t)$ by $\tilde{r}(t)$ and $\tilde{\gamma}(t),$ the system defined by equations (\ref{discrete_s2}), (\ref{discrete_i2}), (\ref{discrete_r2}) and (\ref{discrete_d2}) becomes equivalent to  
\begin{align}
	s(t+1)-s(t)=&\, -\beta(t) s(t)i(t), \nonumber \\
	i(t+1)-i(t)=&\, \beta(t) s(t)i(t)-\tilde{\gamma}(t)i(t), \label{discreteSIR} \\
	\tilde{r}(t+1)-\tilde{r}(t)=&\, \tilde{\gamma}(t)i(t). \nonumber
\end{align}
where $\tilde{\gamma}(t)$ are given by adding (\ref{gammat}) and (\ref{deltat}), that is
\begin{equation}
	\tilde{\gamma}(t)=(\gamma+\delta)(t)=\frac{(r+d)(t+1)-(r+d)(t)}{i(t)},
\end{equation}
or
\begin{equation}\label{gammatil}
	\tilde{\gamma}(t)=\frac{\tilde{r}(t+1)-\tilde{r}(t)}{i(t)}.
\end{equation}
We can rewrite $\beta$ in (\ref{betat2}) as
\begin{equation}\label{betat3}
	\beta(t)=\frac{i(t+1)-i(t)+\tilde{r}(t+1)-\tilde{r}(t)}{i(t)[1-i(t)-\tilde{r}(t)]}
\end{equation}
If we compare the system (\ref{discreteSIR}) with the equations (\ref{discrete_s}), (\ref{discrete_i}), and (\ref{discrete_r}), and if we compare the equations (\ref{gammatil}), and (\ref{betat3}) with (\ref{gammat}), and (\ref{betat}) we will see that they are respectively the same equations which means that the associated models are equivalent. From an algebraic point of view, it means that working with the model that treats the dead and recovered compartments separately is equivalent to working with the model that treats these two compartments together. To this end, we must consider that the recovery rate indicates the individuals in the infected state who recover or die.

\section{Estimating transmission and recovery rates}
\label{estimating}

Following the instructions of Chen \textit{et al.} \cite{CHEN}, we use a common technique for filtering noise in linear systems called \textit{Finite Impulse Response} (FIR) \cite{HG} to track and predict $\beta(t)$ and $\gamma(t)$. First, we must collect data $i(t)$ and $r(t)$ during period $0\leq t\leq T-1$ and measure $\left\{ \beta(t), \gamma(t) \mid 0 \leq t \leq T-2 \right\}$ through equations (\ref{betat}) and (\ref{gammat}). Then we estimate the transmission and recovering rates for period $t\geq T-1$ using 
\begin{equation}\label{betahat}
\hat{\beta}(t) = \, a_0 +\sum_{j = 1}^{J} a_j \beta(t-j),  
\end{equation}
and
\begin{equation}\label{gammahat}
\hat{\gamma}(t) = \, b_0 + \sum_{k = 1}^{K} b_k \gamma(t-k).
\end{equation}
Equations (\ref{betahat}) and (\ref{gammahat}) represent the two FIR filters. The \textit{inputs} of these filters are represented, respectively, by $\beta(t)$ and $\gamma(t).$ The \textit{outputs} $\hat{\beta}(t)$ and $\hat{\gamma}(t)$ of these filters represent the predicted transmission and the predicted recovery rates at time $t$, respectively. The integers $J$ and $K$ are the \textit{orders} of these filters. The numbers $$a_j , \quad j = 0,1,2,\dots , J,$$ and $$b_k , \quad k = 0,1,2, \dots , K$$ are the \textit{coefficients} of these filters. The integers $J$ and $K$ are supposed to satisfy $$0 \,\, <\,\, J,\, K \,\, < \,\, T-2.$$ 

To estimate the coefficients $a_j$ and $b_k$ in their model, Chen  \textit{et al.} \cite{CHEN} chose the \textit{ridge regression} \cite{HASTIE, GHW, KW} method, which is nothing more than an optimization method with regularization to solve the following problems:
\begin{align}
	\arg\!\min_{a_j} \left\{ \sum_{t = J}^{T-2} (\beta(t)-\hat{\beta}(t))^2 + \alpha_1 \sum_{j = 0}^{J} a^2_j \right\}, \label{regprob_a} \\
	\arg\!\min_{b_k} \left\{ \sum_{t = K}^{T-2} (\gamma(t)-\hat{\gamma}(t))^2 + \alpha_2 \sum_{k = 0}^{K} b^2_k \right\}, \label{regprob_b}
\end{align}
where $\alpha_1$ and $\alpha_2$ are the \textit{regularization parameters}. To do this, they used the scikit-learn library (a third-party Python 3 library) to compute the ridge regression. In this article, we also use ridge regression, not through a third-party library, but through our own algorithm built to compute the optimized solution to problems (\ref{regprob_a}) and (\ref{regprob_b}) which is determined by the proposition below. Proposition \ref{prop} establishes the matrix form for the problems (\ref{regprob_a}) and (\ref{regprob_b}), and their solutions. Throughout this paper, all vectors are in column form. We also adopt the same capital letter $T$ to designate the transpose of vectors and matrices.
\newline
\begin{proposition} \label{prop}
Let $\left\{ \beta(t), \gamma(t) \mid 0 \leq t \leq T-2 \right\}$ be the set of data representing the transmission and recovery rates obtained by equations (\ref{betat}) and (\ref{gammat}). Then the problems (\ref{regprob_a}) and (\ref{regprob_b}) are equivalent to
\begin{equation}
	\arg\!\min_{a}
	\begin{Vmatrix} y - X a
	\end{Vmatrix}^2 + \alpha_1
	\begin{Vmatrix} a
	\end{Vmatrix}^2 , \label{matrixregprob_a}
\end{equation}
and
\begin{equation}
	\arg\!\min_{b}
	\begin{Vmatrix} z - Y b
	\end{Vmatrix}^2 + \alpha_2
	\begin{Vmatrix} b
	\end{Vmatrix}^2, \label{matrixregprob_b}
\end{equation}
where $y \in \mathbb{R}^{(T-J-1)}, a \in \mathbb{R}^{(J+1)}, z \in \mathbb{R}^{(T-K-1))},$ and $ b\in \mathbb{R}^{(K+1)}$ are vectors given by
\begin{equation} \label{vecy}
   y=(\beta(J),\beta(J+1),\dots ,\beta(T-3),\beta(T-2))^T,
\end{equation}
\begin{equation} \label{veca}
	a = (a_0, a_1, \dots , a_J)^T,
\end{equation}  
\begin{equation} \label{vecz}
z = (\gamma(K), \gamma(K+1), \dots , \gamma(T-3), \gamma(T-2))^T,
\end{equation} and
\begin{equation} \label{vecb} b=(b_0,b_1, \dots, b_K)^T,
\end{equation} and 
$ X \in \mathbb{R}^{(T-J-1) \times (J+1)}$ and $ \quad  Y \in \mathbb{R}^{(T-K-1) \times (K+1)}$ are matrices whose entries are given by
\begin{align} \label{matX}
X(m,1) & =  1, \\
X(m,n) & =  \beta(J-n+m), \label{matX.2}
\end{align}
for $1\leq m \leq T-J-1,~~ 2\leq n \leq J+1, $ and
\begin{align} \label{matY}
Y(p,1) & =  1, \\
Y(p,q) & =  \gamma(K-q+p), \label{matY.2}
\end{align}
for $1\leq p \leq T-K-1,~~ 2\leq q \leq K+1.$ 

Furthermore, the solution $a^{ridge}$ to the problem (\ref{matrixregprob_a}) is given by
\begin{equation}
	a^{ridge} = \sum_{j = 1}^r \dfrac{\sigma_j (u_j^T y)}{\sigma_j^2 + \alpha_1}v_j. \label{aridge} 
\end{equation}
where the integer $r$ is the rank of the matrix $X,$ the $\sigma_j$ are the singular values of $X,$ the $u_j$ are the left singular vectors of $X,$ and the $v_j$ are the right singular vectors of $X$. \newline
Similarly, we obtain the solution to the problem (\ref{matrixregprob_b}) through the analogous formula
\begin{equation}
	b^{ridge} = \sum_{j = 1}^{\tilde{r}} \dfrac{\tilde{\sigma}_j (\tilde{u}_j^T z)}{\tilde{\sigma}_j^2 + \alpha_2}\tilde{v}_j. \label{bridge} 
\end{equation}
where the $\tilde{r}, \tilde{\sigma}_j, \tilde{u}_j,$ and $\tilde{v}_j,$ are the respective elements obtained by the singular value decomposition theorem applied to the matrix $Y.$
\end{proposition}

\proof First we must prove the equivalence between (\ref{regprob_a}) and (\ref{matrixregprob_a}) (the proof of the equivalence between (\ref{regprob_b}) and (\ref{matrixregprob_b}) is entirely analogous). 

Replacing $\hat{\beta}(t)$ in (\ref{regprob_a}) with the sum in (\ref{betahat}) provides
\begin{align}
	\arg\!\min_{a_j} \left\{ \sum_{t = J}^{T-2} \left[\beta(t)-a_0 - \sum_{j = 1}^{J} a_j \beta(t-j)\right]^2 + \alpha_1 \sum_{j = 0}^{J} a^2_j \right\}. \label{aux1}
\end{align}
Developing the first sum that is inside the braces in $(\ref{aux1})$ yields
\[
	\begin{array}{l}
	\sum_{t = J}^{T-2} \left[\beta(t)-a_0 - \sum_{j = 1}^{J} a_j \beta(t-j)\right]^2  = \\ \\
     = \left\|
        \begin{pmatrix}
	        \beta(J)-a_0 - \sum_{j = 1}^{J} a_j \beta(J-j) \\
	        \beta(J+1)-a_0 - \sum_{j = 1}^{J} a_j \beta(J+1-j) \\
	        \vdots \\
	        \beta(T-2)-a_0 - \sum_{j = 1}^{J} a_j \beta(T-2-j)
       \end{pmatrix}
      \right\|^2   \\ \\
    	= \left\| 
    	\begin{pmatrix}
    		\beta(J) \\
    		\beta(J+1) \\
    		\vdots \\
    		\beta(T-2)
    	\end{pmatrix} - a_0 \begin{pmatrix}
    	1 \\
    	1 \\
    	\vdots \\
    	1
    	\end{pmatrix} - a_1 \begin{pmatrix}
    	\beta(J-1) \\
    	\beta(J) \\
    	\vdots \\
    	\beta(T-3) 
    	\end{pmatrix} - \hdots \right. \\ \\
        \left. \hdots - a_J \begin{pmatrix}
    	\beta(0) \\
    	\beta(1) \\
    	\vdots \\
    	\beta(T-J-2) 
    	\end{pmatrix}
    	\right\|^2  \\ \\
    	= \left\| 
    	\begin{pmatrix}
    		\beta(J) \\
    		\beta(J+1) \\
    		\vdots \\
    		\beta(T-2)
    	\end{pmatrix} \right. - \\ \\
         \left.  \begin{pmatrix}
    	1 & \beta(J-1) & \beta(J-2) &\cdots & \beta(0) \\
    	1 & \beta(J) & \beta(J-1) &\cdots & \beta(1) \\
    	\vdots & \vdots &\vdots &\ddots & \vdots\\
    	1 & \beta(T-4) & \beta(T-5) &\cdots & \beta(T-J-3) \\ 
    	1 & \beta(T-3) & \beta(T-4) &\cdots & \beta(T-J-2)
    	\end{pmatrix} \begin{pmatrix}
    	a_0 \\ a_1 \\ \vdots \\ a_J
    	\end{pmatrix}    	
    	\right\|^2 \\ \\
    	= \begin{Vmatrix} y - X a
    	\end{Vmatrix}^2 ,
     	\end{array}     	 
\]
where $y, X$ and $a$ are respectively given as in (\ref{vecy}), (\ref{matX}), (\ref{matX.2}), and (\ref{veca}).
Since the third sum inside the braces in $(\ref{aux1})$ is trivially equal to $ \| a \|^2,$ it follows that (\ref{regprob_a}) and (\ref{matrixregprob_a}) are equivalent.

Let us now prove the formulas (\ref{aridge}) and (\ref{bridge}). Our proof is based on \cite{HASTIE} and \cite{GVL}. In fact they are consequences of the solution of the general case  
\begin{equation}
	\arg\!\min_{c}
	\| y - Xc \|^2 + \lambda
	\| c \|^2 , \label{MinQLR}
\end{equation}
where $X \in \mathbb{R}^{M \times N}$, $y \in \mathbb{R}^{M \times 1}$, $c \in \mathbb{R}^{N \times 1},$ and $\lambda > 0$ is the regularization parameter. 

Since 
\begin{align*}
	\| y - Xc \|^2 + \lambda
	\| c \|^2 = & \left\| 
		\begin{pmatrix}
			Xc - y \\
			\sqrt{\lambda} c
		\end{pmatrix} \right\|^2 \\ 
        = & \left\| 
		\begin{pmatrix}
			X \\
			\sqrt{\lambda}I
		\end{pmatrix} c -
		\begin{pmatrix}
			y \\
			0
		\end{pmatrix}
	\right\|^2 ,
\end{align*}
then the solution of (\ref{MinQLR}) is equivalent to the solution of the normal equation 
\begin{equation*}
	\begin{pmatrix}
		X \\
		\sqrt{\lambda}I
	\end{pmatrix}^T
	\begin{pmatrix}
		X \\
		\sqrt{\lambda}I
	\end{pmatrix}c = 
	\begin{pmatrix}
		X \\
		\sqrt{\lambda}I
	\end{pmatrix}^T 
	\begin{pmatrix}
		y \\
		0
	\end{pmatrix},
\end{equation*}
or, equivalently
\begin{equation} \label{normalequation}
	(X^TX + \lambda I) c = X^T y,
\end{equation}
where $I$ is the $N\times N$ identity matrix.

According to {\it Singular Value Decomposition} theorem \cite{GVL,HASTIE} there exist orthogonal matrices $U = [u_1, \dots, u_M] \in \mathbb{R}^{M \times M},$ $V = [v_1, \dots, v_N] \in \mathbb{R}^{N \times N},$ and numbers 
$$\sigma_1 \ge \dots \ge \sigma_r > 0, \hspace{0.5cm} r \leq \min(M, N),$$
such that $U^T XV$ is the block diagonal matrix
\begin{equation} \label{svd}
	U^T X V = D = \left(\begin{array}{cc}
		\Sigma_r & 0 \\
		0 & 0 
	\end{array}\right) \in \mathbb{R}^{M\times N},     
\end{equation}
where $\Sigma_r = {\rm diag}(\sigma_1, \sigma_2, ..., \sigma_r) \in \mathbb{R}^{r \times r}.$ 
The integer $r$ is the rank of $X,$ the numbers $\sigma_1, \sigma_2, \cdots, \sigma_r$ are the {\it singular values} of $X,$ the column vectors $u_j$ of $U$ are the {\it left singular vectors} of $X,$ and the column vectors $v_j$ of $V$ are the {\it right singular vectors} of $X.$ 

Replacing $X$ in equation (\ref{normalequation}) with your singular value decomposition $UDV^T$ produces
\begin{equation}\label{normalequation2}
	(D^TD + \lambda I)\xi = D^TU^T y,
\end{equation}
where $\xi = V^Tc.$ After using (\ref{svd}) in (\ref{normalequation2}) we realize that  
$D^TD + \lambda I$ is the diagonal matrix $${\rm diag}(\sigma_1^2 + \lambda, \dots, \sigma_r^2 + \lambda, \lambda, \dots, \lambda) \in \mathrm{R}^{N \times N},$$ and $D^TU^T$ is the matrix whose rows are given by the vectors 
$$\begin{cases}
	\sigma_j u^T_j, & j = 1, \cdots,r \\
	0, & j = r+1, \cdots, N.
\end{cases}$$
Thus, the solution to the matrix equation (\ref{normalequation2}) is given by
\begin{equation}\label{solNormal}
	\xi_j = \begin{cases}
		\dfrac{\sigma_j (u^T_j y)}{\sigma^2_j + \lambda}, & j = 1 : r \\
		0, & j = r+1 : n.
	\end{cases}
\end{equation}
Then the solution to (\ref{normalequation}) is obtained by
\begin{align*}
	c = V\xi = & \begin{bmatrix} v_1 & \cdots & v_r & v_{r+1} & \cdots & v_N\end{bmatrix} \begin{bmatrix}
		\xi_1 \\
		\vdots \\
		\xi_r \\
		0 \\
		\vdots \\
		0
	\end{bmatrix} \\ 
    = & \sum_{j = 1}^r \xi_j v_j .
\end{align*}
Replacing $\xi_j$ in the expression above with the values found in (\ref{solNormal}), we have that the solution to the regularized linear least squares problem (\ref{MinQLR}) is given by:
\begin{equation}\label{SolRidge}
	c^{ridge} = \sum_{j = 1}^r \dfrac{\sigma_j (u_j^T y)}{\sigma_j^2 + \lambda}v_j. \hspace{0.5in}\rule{2mm}{2mm}
\end{equation}
The components of the vector $c^{ridge}$ in (\ref{SolRidge}) are the real numbers that minimize the objective function of (\ref{MinQLR}) and will be called {\it ridge coefficients.}

Since
\begin{equation}\label{limRidge}
	\lim_{\lambda \rightarrow 0} c^{ridge} = \lim_{\lambda \rightarrow 0} \sum_{j = 1}^r \dfrac{\sigma_j (u_j^T y)}{\sigma_j^2 + \lambda}v_j = \sum_{j = 1}^r \dfrac{(u_j^T y)}{\sigma_j}v_j,
\end{equation}
we see that the sum on the right-hand side of (\ref{limRidge}) represents the solution to the non-regularized problem
\begin{equation}\label{MQLnaoReg}
	\arg\!\min_{c}
	\begin{Vmatrix} y - Xc
	\end{Vmatrix}_2^2,
\end{equation}
that is, the solution of the regularized problem (\ref{MinQLR}) converges to the solution of the non-regularized linear least squares problem (\ref{MQLnaoReg}) when $\lambda \rightarrow 0$.

If we compare the sums on the right-hand side of both expressions (\ref{SolRidge}) and (\ref{limRidge}) we will see that the addition of the regularization portion $\lambda \| c \|_2^2$ to the linear least squares problem acts like a filter. In fact, analyzing the general term of the sum in (\ref{SolRidge}) we see that: \textit{(i)} if $ \lambda \ll \sigma_j,$ then 
\begin{equation*}
    \dfrac{\sigma_j (u_j^T y)}{\sigma_j^2 + \lambda} \approx \dfrac{u_j^T y}{\sigma_j},
\end{equation*}
that is, the portions of the sum corresponding to singular values that are much larger than the regularization parameter $\lambda$ remain almost unchanged when they are compared with the respective portions of the sum from the solution of the non-regularized problem. On the other hand, \textit{(ii)} if $ \lambda \gg \sigma_j,$ then $ \sigma_j^2 + \lambda \neq 0,$ and $ \sigma_j \approx 0,$ which means
\begin{equation*}
    \dfrac{\sigma_j (u_j^T y)}{\sigma_j^2 + \lambda} \approx 0,
\end{equation*}
i.e., the portions corresponding to singular values that are much smaller than $\lambda$ are practically zero.

Let us now display a version of the algorithm of Chen  \textit{et al.} \cite{CHEN} adapted to the use of variables $s, i$ and $r$ instead of $S, I$ and $R,$ and considering $s(t)\not\approx 1$ (equivalently $S(t)\not\approx n$). This algorithm is useful for finding the estimates $\beta(t)$ and $\gamma(t)$ at time $t = T-1,$ and $i(t)$ and $r(t)$ at $t = T,$ assuming that the data $\{i(0), i(1), \cdots, i(T-1)\}, \{r(0), r(1), \cdots, r(T-1)\},$  
$ \{\beta(0), \beta(1), \cdots, \beta(T-2)\},$ and $ \{\gamma(0), \gamma(1), \cdots, \gamma(T-2)\}  $
are known.
 
The estimates $\beta(t)$ and $\gamma(t)$ in $t = T-1$ are given by $\hat{\beta}(T-1)$ and $\hat{\gamma}( T-1)$ whose values are obtained by (\ref{betahat}) and (\ref{gammahat}) after replacing $t$ by $T-1,$ and the coefficients $a_j$ and $b_k,$ respectively, by the components of the vectors $a^{ridge} $ and $ b^{ridge}.$

Making $t=T-1$ in both (\ref{i_predicted}) and (\ref{r_predicted}) we obtain the predictions of $i$ and $r$ at time $t=T,$ named by $\hat{\textrm{\i}}(T)$ and $\hat{r}(T),$ so that
\begin{align}
	\hat{\textrm{\i}}(T) = &\,\, [1 + \hat{\beta}(T-1)(1-i(T-1)-r(T-1))- \nonumber \\
    & \,\, \hat{\gamma}(T-1)]\,\, i(T-1) \label{ihat} \\
	\hat{r}(T) = &\,\, r(T-1) + \hat{\gamma}(T-1)i(T-1). \label{rhat}
\end{align}

\begin{algorithm}
\SetAlgoLined
	\caption{Prediction algorithm for the discrete Time-dependent SIR model adapted to the variables $s(t),i(t),$ and $r(t),$ and with the assumption $s(t) \neq 1, \,\,\, t\,\geq \, 0$.} 
    \label{alg1}
	\KwIn{$\{i(t), r(t), 0 \le t \le T-1\}$, regularization parameters $\alpha_1$ and $\alpha_2$, order of FIR filters $J$ and $K,$ and prediction window $W.$ \par}
	\KwOut{$\{\beta(t), \gamma(t), \,\,\,  0 \le t \le T-2\}, \newline 
     \{\hat{\beta}(t), \hat{\gamma}(t), \,\,\, T-1 \le t \le T+W-2\},$ and \newline  $\{\hat{I}(t), \hat{R}(t),\,\,\,  T \le t \le T+W-1\}$.}
 \BlankLine
    \KwSteps{
    
    \nl Measure $\{\beta(t), \gamma(t), 0 \le t \le T-2\}$, using (\ref{betat}) and (\ref{gammat}) respectively. \par
    \nl With the filters $J$ and $K,$ and the data obtained from step 1, calculate $y, X, z$ and $Y$ using (\ref{vecy}),(\ref{matX}), (\ref{matX.2}), (\ref{vecz}), (\ref{matY}), and (\ref{matY.2}). \par
    \nl Calculate the vectors $a^{ridge}$ and $b^{ridge}$ as in (\ref{aridge}) and (\ref{bridge}). \par
    \nl Exchange the coefficients $a_j$ and $b_k$ in (\ref{betahat}) and (\ref{gammahat}) for the components of the vectors $a^{ridge}$ and $b^{ridge}$ respectively and then calculate the estimates $\hat{\beta}(T-1)$ and $\hat{\gamma}(T-1).$ \par
    \nl Estimate the fractions of infected persons $\hat{\textrm{\i}}(T)$ and recovered persons $\hat{r}(T)$ on day $T$ using (\ref{ihat}) and (\ref{rhat}). \par	 
    \nl \While{$ T \leq t \leq T+W-2$}{ \nl	Estimate $\hat{\beta}(t)$ and $\hat{\gamma}(t)$ in (\ref{betahat}) and (\ref{gammahat}) respectively as in steps 4. and 5. \par
    \nl Predict $\hat{\textrm{\i}}(t+1)$ and $\hat{r}(t+1)$ using (\ref{ihat2}) and (\ref{rhat2}).} 
    \nl Calculate the predicted values of the number of infected persons $\hat{I}(t)$ and recovered persons $\hat{R} (t)$ for $T \leq t \leq T+W-1$ using (\ref{fractions})}
\end{algorithm}

Finally, to obtain the other predictions $\hat{\textrm{\i}}(t)$ and $\hat{r}(t)$, for $t = T+1, T+2, \dots, T+W$, where $ W$ represents, as we said before, a prediction window, we must first obtain the estimates $\hat{\beta}(t)$ and $\hat{\gamma}(t)$, for $t = T, T+1 , \cdots, T+W-1$, whose values are obtained by setting $t = T, T+1, \cdots, T+W-1$ in equations (\ref{betahat}) and (\ref{gammahat}), after replacing the coefficients $a_j$ and $b_k$ in these respective equations with the coefficients of the vectors $a^{ridge}$ and $b^{ridge}$. Then, as in (\ref{ihat}) and (\ref{rhat}), we obtain the estimates:
\begin{align}
	\hat{\textrm{\i}}(t+1) &= [1 + \hat{\beta}(t)(1-\hat{\textrm{\i}}(t)-\hat{r}(t))- \hat{\gamma}(t)]\, \hat{\textrm{\i}}(t), \label{ihat2} \\
	\hat{r}(t+1) &= \hat{r}(t) + \hat{\gamma}(t)\, \hat{\textrm{\i}}(t).\label{rhat2}
\end{align}
for all $t \ge T.$
The detailed steps of the method are described in Algorithm \ref{alg1}, which is based on the algorithm of Chen \textit{et al.} \cite{CHEN} with adaptations.

\section{A modification to Algorithm 1.}
\label{modification}

We would now like to propose a small modification to the method in Algorithm \ref{alg1}.

Like the algorithm proposed by Chen \textit{et al.} \cite{CHEN}, Algorithm \ref{alg1} uses the same coordinates of the vectors $a^{ridge}$ and $b^{ridge}$ obtained in step 3 to obtain in step 6 the estimates $\hat{\beta}(t)$ and $\hat{\gamma}(t)$ when $T \leq t \leq T+W-1.$

Our proposal is to update the vectors $a^{ridge}$ and $b^{ridge}$ at each step we want to calculate the next estimate of $\hat{\beta}$ and $\hat{\gamma}.$ We do this by updating the variable $T$ through the command $T \gets T+1$ and then calculating the new vectors $a^{ridge}$ and $b^{ridge}.$ at each step. 

To facilitate the explanation of our modification mentioned in the previous paragraph, we must first provide the vector version that determines the elements of Algorithm \ref{alg1}, starting with the values of $\hat{\beta}(t)$ and $\hat{\gamma}(t)$ that should be placed in the form of the Euclidean inner product (defined by $<x,y>\, = \, x^Ty $) as detailed below.
%
Let us denote by $ \hat{a}_0, \hat{a}_1, \dots, \hat{a}_J $
the components of the vector $a^{ridge}$ which as we know is the solution of the regularized least squares problem (\ref{matrixregprob_a}), i.e.,
\begin{equation} \label{aridge_v2}
	a^{ridge}\, =\, \arg\!\min_{a}
	\begin{Vmatrix} y - X a
	\end{Vmatrix}^2 + \alpha_1
	\begin{Vmatrix} a
	\end{Vmatrix}^2 ,
\end{equation} 
where the vector $y$ and the matrix $X$ are built as in (\ref{vecy}), (\ref{matX}), and (\ref{matX.2}) from the known data vector
\begin{equation} \label{beta0}
    \boldsymbol{\beta} := (\beta(0),\beta(1), \dots, \beta(T-2))^T.
\end{equation} 
As we already know the coordinates of the vector $\boldsymbol{\beta}$ are the data of the parameter $\beta(t)$ collected from the instant $t=0$ to the instant $t=T-2,$ which means that 
\begin{equation} \label{betacoord}
  \boldsymbol{\beta}(m) = \beta(m-1), \quad m=1,2,\dots, T-1.  
\end{equation}
It follows from step 4 of the Algorithm \ref{alg1} that
\begin{align} \label{aux2}
\hat{\beta}(T-1) = \begin{bmatrix} \hat{a}_0 & \hat{a}_1 & \hat{a}_2 &\cdots & \hat{a}_J\end{bmatrix} \begin{bmatrix}
		1 \\
		\beta(T-2) \\
		\beta(T-3) \\
		\vdots \\
		\beta(T-J-1)
	\end{bmatrix}.
\end{align}
Note that the column vector on the right side of (\ref{aux2}) represents the vector of $\mathbb{R}^{J+1}$ whose first component is 1 and the other components are the last $J$ coordinates of $\boldsymbol{\beta}$. We will denote such a vector by $\boldsymbol{\beta_J},$ i.e.,
\begin{equation} \label{betaJ}
    \boldsymbol{\beta_J} = (1,\beta(T-2), \beta(T-3), \dots, \beta(T-J-1))^ T.
\end{equation}
Then the equation (\ref{aux2}) can be rewritten as
\begin{equation} \label{aux3}
    \hat{\beta}(T-1) \, \,  = \quad < a^{ridge}, \boldsymbol{\beta_J} >. 
\end{equation}

The next step is to represent $\hat{\beta}(T)$ as an inner product. To do this we must first add to the vector $\boldsymbol{\beta}$ an additional component given by the estimate $\hat{\beta}(T-1)$ such that we have defined the following next vector
\begin{align} \label{beta1}
\boldsymbol{\beta^{(1)}} := & \,\, (\boldsymbol{\beta}^{ T},\hat{\beta}(T-1))^T \nonumber \\
= &\,\,  (\beta(0),\beta(1), \dots, \beta(T-2), \hat{\beta}(T-1))^T. 
\end{align}
Similarly we define $\boldsymbol{\beta^{(1)}_J}$ the column vector of $\mathbb{R}^{J+1}$ whose first component is 1 and the remaining components are the last $J$ coordinates of $\boldsymbol{\beta^{(1)}},$ i.e.,
\begin{equation} \label{beta1J}
  \boldsymbol{\beta^{(1)}_J} := (1,\hat{\beta}(T-1),\beta(T- 2), \beta(T-3), \dots, \beta(T-J))^T.   
\end{equation}
Then, according to Algorithm \ref{alg1} we have 
\begin{equation} \label{aux4}
 \hat{\beta}(T) \, \,  = \quad < a^{ridge}, \boldsymbol{\beta^{(1)}_J} >.   
\end{equation}

To represent the next estimate $\hat{\beta}(T+1)$ we must proceed in a manner entirely analogous to the two previous steps: we first define the vectors 
\begin{align}\label{beta2}
  \boldsymbol{\beta^{(2)}} := &\, \,((\boldsymbol{\beta^{(1)}})^T, \hat{\beta}(T))^T \\
        = & \,\, (\beta(0),\beta(1), \dots, \beta(T-2), \hat{\beta}(T-1), \hat{\beta}(T))^T.  \nonumber
\end{align}
and 
\begin{align*}\boldsymbol{\beta^{(2)}_J} := & \,\,(1,\hat{\beta}(T),\hat{\beta}(T-1),\beta(T-2), \beta(T-3), \dots \\
    &\,\,\dots,\beta(T-J+1))^T, 
\end{align*} 
and then we calculate
\begin{equation*}   \hat{\beta}(T+1) \, \,  = \quad < a^{ridge}, \boldsymbol{\beta^{(2)}_J} >. 
\end{equation*}

Continuing this inductive procedure we will be able to establish all inner product formulas that determine the estimated parameters $\hat{\beta}(t)$  established by Algorithm \ref{alg1}:
$$  \hat{\beta}(T+j-1) \, \,  = \quad < a^{ridge}, \boldsymbol{\beta^{(j)}_J} >, $$
where
\begin{align} \label{betaJj}
  \boldsymbol{\beta^{(j)}_J} \,\,\, := &\,\, (1,\hat{\beta}(T+j-2),\hat{\beta}(T+j-3),\dots \\
  & \dots, \hat{\beta}(T-1), \beta(T-2), \dots, \beta(T-J+j-1))^T,  \nonumber 
\end{align}
for each $j=1,2,\dots,W.$

By an entirely analogous procedure we can obtain similar vector and inner product formulas for the predicted parameters of recovered persons $\hat{\gamma}(t),$ i.e.,
\begin{equation} \label{aux5}
 \hat{\gamma}(T-1) \, \,  = \, < b^{ridge}, \boldsymbol{\gamma_K} >,
\end{equation}
where
\begin{equation} \label{bridge_v2}
	b^{ridge}\, =\, \arg\!\min_{b}
	\begin{Vmatrix} z - Y b
	\end{Vmatrix}^2 + \alpha_2
	\begin{Vmatrix} b
	\end{Vmatrix}^2 ,
\end{equation} 
\begin{equation} \label{gammaK}
\boldsymbol{\gamma_K} \,\, = (1,\gamma(T-2), \gamma(T-3), \dots, \gamma(T-K-1))^ T,    
\end{equation}
and, for each $j = 1, 2, \cdots, W,$
$$ \hat{\gamma}(T+j-1) \, \,  = \quad < b^{ridge}, \boldsymbol{\gamma^{(j)}_K} >\, , $$
\begin{align} \label{gammaKj}
 \boldsymbol{\gamma^{(j)}_K} \,\,\, := &\,\, (1,\hat{\gamma}(T+j-2),\hat{\gamma}(T+j-3),\dots \\
 &\dots, \hat{\gamma}(T-1), \gamma(T-2), \dots \gamma(T-K+j-1))^T. \nonumber
\end{align}
\subsection{Explanation of method modification} 

The central idea of the method we will propose is summarized as follows: we use the current estimate $\hat{\beta}(t), \, \, t \geq T-1,$ together with the previous transmission rates to construct a new vector $y $ and a new matrix $X,$ then solve the regularized linear least squares problem associated with these new elements, and finally determine the estimate at the next day $t+1.$ 
We will illustrate this method with the first three steps to obtain the estimates $\hat{\beta}(T-1), \hat{\beta}(T)$ and $\hat{\beta}(T+1).$ The remaining estimates can be obtained inductively.

The objective of our first step is to obtain the estimate $\hat{\beta}(T-1)$ whose value must be calculated the same way as in (\ref{aux3}),
$$  \hat{\beta}(T-1) \, \,  = \quad < a^{ridge}, \boldsymbol{\beta_J} >.$$

Rewriting both the coordinates of the vector $y$ given as in (\ref{vecy}) and the entries of the matrix $X$ given as in (\ref{matX}) and (\ref{matX.2}) in terms of the components of $\boldsymbol{\beta}$ as in (\ref{betacoord}), we have
\begin{equation} \label{vecy_2}
   y(m) = \boldsymbol{\beta}(J+m), \quad \forall \,\,\,\, 1\leq m \leq T-J-1, 
\end{equation}  and
\begin{align*}
X(m,1) & =  1, \\
X(m,n) & = \boldsymbol{\beta}(J-n+m+1),
\end{align*}
for $1\leq m \leq T-J-1,~~ 2\leq n \leq J+1, $

In the second step, we first define the new vector $y^{(1) } \in \mathbb{R}^{(T_1-(J+1))}$ and the new matrix $X^{(1)} \in \mathbb{R}^{(T_1-J-1) \times (J+1)},$ with $T_1 = T+1,$ using the coordinates of the vector $\boldsymbol{\beta^{(1) }}$ in (\ref{beta1}), i.e.,
\begin{equation} \label{vecy1}
y^{(1)}(m) = \boldsymbol{\beta^{(1)}}(J+m),\quad \forall \,\,\,\, 1\leq m \leq T-J, \nonumber
\end{equation}
and
\begin{align*} 
X^{(1)}(m,1) & =  1, \\
X^{(1)}(m,n) & = \boldsymbol{\beta^{(1)}}(J-n+m+1),
\end{align*}
for $1\leq m \leq T-J,~~ 2\leq n \leq J+1. $

We now solve the new regularized linear least squares problem
\begin{equation}
	\arg\!\min_{a}
	\begin{Vmatrix} y^{(1)} - X^{(1)} a
	\end{Vmatrix}^2 + \alpha_1
	\begin{Vmatrix} a
	\end{Vmatrix}^2 , \nonumber 
\end{equation}
whose solution $ a^{ridge,1}$ we can obtain by using the proposition \ref{prop}. 
With $ a^{ridge,1}$ and the vector $\boldsymbol{\beta^{(1)}_J}$ in (\ref{beta1J}) we calculate $\hat{\beta}(T)$ in a similar way to (\ref{aux4}), i.e.,  
\begin{equation} \label{aux6}
 \hat{\beta}(T) \, \,  = \quad < a^{ridge,1}, \boldsymbol{\beta^{(1)}_J} >.  \nonumber 
\end{equation}

In the third step, we follow the previous step procedures to first determine the entries of the next new vector $y^{(2)} \in \mathbb{R}^{(T_2-(J+1))}$ and the next new matrix $X^{(2)} \in \mathbb{R}^{(T_2-J-1) \times (J+1)},$ where $T_2 = T+2,$ by using the coordinates of the vector $\boldsymbol{\beta^{(2)}}$ in (\ref{beta2}), so that 
$$ y^{(2)}(m) = \boldsymbol{\beta^{(2)}}(J+m),\quad \forall \,\,\,\, 1\leq m \leq T-J+1,$$
and
\begin{align*} 
X^{(2)}(m,1) & =  1, \\
X^{(2)}(m,n) & = \boldsymbol{\beta^{(2)}}(J-n+m+1),
\end{align*}
for $1\leq m \leq T-J+1,~~ 2\leq n \leq J+1. $

We now use the solution
\begin{equation}
	a^{ridge,2} = \arg\!\min_{a}
	\begin{Vmatrix} y^{(2)} - X^{(2)} a
	\end{Vmatrix}^2 + \alpha_1
	\begin{Vmatrix} a
	\end{Vmatrix}^2 \nonumber
\end{equation}
to calculate the next estimate
$$ \hat{\beta}(T+1) \, \,  = \quad < a^{ridge,2}, \boldsymbol{\beta^{(2)}_J} >. $$

We have inductively for each $j = 1, 2,\cdots, W-1:$
\begin{equation} \label{aux7}
 \hat{\beta}(T+j-1) \, \,  = \quad < a^{ridge,j}, \boldsymbol{\beta^{(j)}_J} >, 
\quad  \mbox{$\boldsymbol{\beta^{(j)}_J}$ as in (\ref{betaJj}),}
\end{equation}
\begin{equation} \label{aridgej}
	a^{ridge,j} = \arg\!\min_{a}
	\begin{Vmatrix} y^{(j)} - X^{(j)} a
	\end{Vmatrix}^2 + \alpha_1
	\begin{Vmatrix} a
	\end{Vmatrix}^2,
\end{equation}
\begin{equation} \label{vecyj}
y^{(j)}(m) = \boldsymbol{\beta^{(j)}}(J+m),\quad \forall \,\,\,\, 1\leq m \leq T-J+j-1,
\end{equation}
and
\begin{align} 
X^{(j)}(m,1) & =  1, \label{matXj} \\
X^{(j)}(m,n) & = \boldsymbol{\beta^{(j)}}(J-n+m+1), \label{matXj.2}
\end{align}
for $1\leq m \leq T-J+j-1,~~ 2\leq n \leq J+1, $
where $\boldsymbol{\beta^{(j)}}$ is the vector of $\mathbb{R}^{(T+j-1)}$ given by
\begin{align} 
  \boldsymbol{\beta^{(j)}} = & \, \, (\beta(0),\beta(1), \dots, \beta(T-2), \hat{\beta}(T-1), \hat{\beta}(T), \dots \nonumber \\
  &\dots, \hat{\beta}(T+j-2) )^T.  \label{beta_j}
\end{align}
By a procedure entirely analogous, we can obtain, for each $j = 1, 2,\cdots, W-1,$ the equivalent expressions to find the estimates $\hat{\gamma}(t)$ as follows:
\begin{equation} \label{aux8}
   \hat{\gamma}(T+j-1) \, \,  = \quad < b^{ridge,j}, \boldsymbol{\gamma^{(j)}_K} >\, , \quad \mbox{$\boldsymbol{\gamma^{(j)}_K}$ as in (\ref{gammaKj}),}  
\end{equation}
\begin{equation} \label{bridgej}
	b^{ridge,j} = \arg\!\min_{b}
	\begin{Vmatrix} z^{(j)} - Y^{(j)} b
	\end{Vmatrix}^2 + \alpha_2
	\begin{Vmatrix} b
	\end{Vmatrix}^2,
\end{equation}
\begin{equation} \label{veczj}
z^{(j)}(m) = \boldsymbol{\gamma^{(j)}}(K+m),\quad \forall \,\,\,\, 1\leq m \leq T-K+j-1,
\end{equation}
and
\begin{align} 
Y^{(j)}(m,1) & =  1, \label{matYj} \\
Y^{(j)}(m,n) & = \boldsymbol{\gamma^{(j)}}(K-n+m+1), \label{matYj.2}
\end{align}
for $1\leq m \leq T-K+j-1,~~ 2\leq n \leq K+1, $
where $\boldsymbol{\gamma^{(j)}}$ is the vector of $\mathbb{R}^{(T+j-1)}$ given by
\begin{align}\label{gamma_j} 
  \boldsymbol{\gamma^{(j)}} = & \,\, (\gamma(0),\gamma(1), \dots, \gamma(T-2), \hat{\gamma}(T-1), \hat{\gamma}(T), \dots \nonumber\\
  &\dots, \,\, \hat{\gamma}(T+j-2) )^T.  
\end{align}

Next, we present the method modification discussed above in algorithm format. The first five steps are the same as the Algorithm \ref{alg1}. The modification is described from step 6 to step 11. 
\begin{algorithm}
\SetAlgoLined
	\caption{Prediction algorithm for the discrete Time-dependent SIR model with modifications (Modified Algorithm \ref{alg1}).}
    \label{alg2}
	\KwIn{$\{i(t), r(t), 0 \le t \le T-1\}$, regularization parameters $\alpha_1$ and $\alpha_2$, order of FIR filters $J$ and $K,$ and prediction window $W.$ \\}
	\KwOut{$\{\beta(t), \gamma(t), \,\,\, 0 \le t \le T-2\}, \newline \{\hat{\beta}(t), \hat{\gamma}(t), \,\,\, T-1 \le t \le T+W-2\},$ and \newline $\{\hat{I}(t), \hat{R}(t), \,\,\, T \le t \le T+W-1\}$.}
 \BlankLine
    \KwSteps{    
    
    \nl Measure $\{\beta(t), \gamma(t), 0 \le t \le T-2\}$, using (\ref{betat}) and (\ref{gammat}) respectively.\\ 
    \nl With the filters $J$ and $K,$ and the data obtained from step 1, calculate $y, X, z$ and $Y$ using (\ref{vecy}),(\ref{matX}), (\ref{matX.2}), (\ref{vecz}), (\ref{matY}), and (\ref{matY.2}). \\
    \nl Calculate the vectors $a^{ridge}$ and $b^{ridge}$ as in (\ref{aridge}) and (\ref{bridge}). \\
    \nl Exchange the coefficients $a_j$ and $b_k$ in (\ref{betahat}) and (\ref{gammahat}) for the components of the vectors $a^{ridge}$ and $b^{ridge}$ respectively and then calculate the estimates $\hat{\beta}(T-1)$ and $\hat{\gamma}(T-1).$ \\
    \nl Estimate the fractions of infected persons $\hat{\textrm{\i}}(T)$ and recovered persons $\hat{r}(T)$ on day $T$ using (\ref{ihat}) and (\ref{rhat}). \\	
    \nl \for{$ j = 1:W-1$}{
    \nl	Calculate the vectors $\boldsymbol{\beta^{(j)}}, \boldsymbol{\beta^{(j)}_J}, \boldsymbol{\gamma^{(j)}},$ and $ \boldsymbol{\gamma^{(j)}_K}$ using (\ref{beta_j}), (\ref{betaJj}), (\ref{gamma_j}) and (\ref{gammaKj}) respectively. \\
    \nl Use (\ref{vecyj}), (\ref{matXj}), (\ref{matXj.2}), (\ref{veczj}), (\ref{matYj}), and (\ref{matYj.2}) to calculate $y^{(j)}, X^{(j)}, z^{(j)},$ and $ Y^{(j)}$ respectively.\\
    \nl Use Prop. \ref{prop} to solve the regularized LS problems (\ref{aridgej}) and (\ref{bridgej}), and calculate the vectors $a^{ridge,j}$ and $b^{ridge,j}.$ \\
    \nl Do $t=T+j-1,$ and calculate $\hat{\beta}(t), $ and $ \hat{\gamma}(t)$ as in (\ref{aux7}) and (\ref{aux8}), \\
    \nl and predict $\hat{\textrm{\i}}(t+1),$ and $\hat{r}(t+1)$ using (\ref{ihat2}) and (\ref{rhat2}).
      }
    \nl Calculate the predicted values of the number of infected persons $\hat{I} (t)$ and recovered persons $\hat{R} (t)$ for $T \leq t \leq T+W-1$   using (\ref{fractions})
    }
\end{algorithm}
\newpage
We conclude this subsection by showing the main differences between the original method by Chen  \textit{et al.} \cite{CHEN} and our modified method. 

The original algorithm proposed by Chen \textit{et al.} \cite{CHEN} can be described by Algorithm \ref{alg1} of this article, as long as we take care to change the formulas that allow the calculation of the parameters $\beta(t)$ and $\gamma(t),$ since they assume $S(t) \approx n$ in their algorithm. We must also adapt the notation of the variables, since they work directly with the numbers of susceptible people $S(t)$, infected $I(t),$ and recovered $R(t)$ instead of their respective fractions (over the total population $n$) $s(t), i(t)$ and $r(t).$

Assuming $S(t) = n$ in the equations (\ref{discretesystem}), the formulas for calculating $\beta(t)$ and $\gamma(t)$ become
\begin{equation}
	\beta(t) = \dfrac{[I(t+1)- I(t)]+[R(t+1)- R(t)]}{I(t)}, \label{chen_betat}
\end{equation}
and
\begin{equation}
	\gamma(t) = \dfrac{R(t+1)- R(t)}{I(t)}, \label{chen_gammat}
\end{equation}
and the formulas (\ref{ihat2}) and (\ref{rhat2}) are changed to
\begin{align}
	\hat{I}(t+1) =&\,\, [1+\hat{\beta}(t) - \hat{\gamma}(t)] \hat{I}(t), \label{I_predicted} \\
	\hat{R}(t+1) =&\,\, \hat{R}(t)+\hat{\gamma}(t) \hat{I}(t). \label{R_predicted}
\end{align} 
Therefore, step 1 in Algorithm \ref{alg1} must be performed by using the equations (\ref{chen_betat}) and (\ref{chen_gammat}). Consequently, the matrices $y, X, z,$ and $Y$ in (\ref{vecy}), (\ref{matX}), (\ref{matX.2}), (\ref{vecz}), (\ref{matY}), and (\ref{matY.2}) mentioned in step 2 will have their entries modified, as well as all elements to be calculated in steps 3, 4, and 7. Step 5 (respectively, step 8) must be performed by using equations (\ref{I_predicted}) and (\ref{R_predicted}) for $t = T-1.$ (respectively $t = T$). 

Table \ref{tab1} lists the comparative differences between the original method by Chen \textit{et al.} \cite{CHEN} and the modified method.
\begin{table}
\caption{\textbf{Comparative differences between the method by Chen \textit{et al.} \cite{CHEN}, and the modified method.}}
\setlength{\tabcolsep}{3pt}
\begin{tabular}{|p{90pt}|p{80pt}|p{85pt}|}
\hline
  &
\textsc{Original Method} &
\textsc{Modified Method} \\
\hline
Model variables. &
$ S(t), I(t), \textrm{and}\, R(t).$ &
$s(t), i(t),$ and $r(t).$ \\
\hline
Assumption. & $S(t) \approx n, \,\, t\, \geq\, 0. $ &
$s(t) \neq 1, \,\, t \geq\, 0. $ \\
\hline
Calculation of $\beta(t)$ and $\gamma(t).$ &
As in (\ref{chen_betat}) and (\ref{chen_gammat}). &
 As in (\ref{betat}) and (\ref{gammat}). \\
 \hline
Calculation of estimates  $\hat{\beta}(T+j-1)$ \par for $1\, \leq \, j\, \leq \, W-1. $     &  $\hat{\beta}(T+j-1) \, \,  =$  \par $ < a^{ridge}, \boldsymbol{\beta^{(j)}_J} >, $ \par $a^{ridge}$ as in (\ref{aridge_v2}). &  $\hat{\beta}(T+j-1) \, \,  =$ \par 
$ < a^{ridge,j}, \boldsymbol{\beta^{(j)}_J} >, $ \par  $a^{ridge,j}$ as in (\ref{aridgej}). \\ 
 \hline
Calculation of estimates $\hat{\gamma}(T+j-1)$ \par for $1\, \leq \, j\, \leq \, W-1. $     &  $\hat{\gamma}(T+j-1) \, \,  =$  \par $ < b^{ridge}, \boldsymbol{\gamma^{(j)}_K} >, $ \par $b^{ridge}$ as in (\ref{bridge_v2}). &  $\hat{\gamma}(T+j-1) \, \,  =$ \par 
$ < b^{ridge,j}, \boldsymbol{\gamma^{(j)}_K} >, $ \par  $b^{ridge,j}$ as in (\ref{bridgej}). \\ 
 \hline
Estimate of Infected \par Persons & $\hat{I}(t+1)$  as in (\ref{I_predicted}), \par  for all $t\ge T-1.$ & $\hat{I}(t+1) = n \, \hat{\textrm{\i}}(t+1),$ \par  $\hat{\textrm{\i}}(t+1)$ as in (\ref{ihat2}), \par for all $t\ge T-1.$ \\
\hline
Estimate of Recovered Persons & $\hat{R}(t+1)$  as in (\ref{R_predicted}), \par  for all $t\ge T-1.$ & $\hat{R}(t+1) = n \, \hat{r}(t+1),$ \par  $\hat{r}(t+1)$ as in (\ref{rhat2}), \par for all $t\ge T-1.$ \\
\hline
\end{tabular}
\label{tab1}
\end{table}
\subsection{Numerical results} 
\label{numericalresults}

In this subsection, we use the modified method described in Algorithm \ref{alg2} to predict the number of infected and recovered persons (in a prediction window) of COVID-19 in the state of Minas Gerais, Brazil, throughout 2020, a period in which vaccination had not yet started in the country\footnote{https://fiocruz.br/noticia/2022/01/vacinacao-contra-covid-19-no-brasil-completa-um-ano}. All numerical implementations were performed in the open source and free Octave software environment. We also compare the results obtained with our modified method with the results obtained by the original method of Chen \textit{et al.} \cite{CHEN}. For our analysis, we collected our data set from the repository available at \cite{COTA} and formed a table with two columns, the first containing the daily accumulated number of infected persons, and the second containing the daily accumulated number of recovered persons added to the accumulated number of deaths from the disease (according to the discussion in section \ref{equivalence}, the number of deaths must be counted in recovery state). These accumulated data refer to the days starting on April 1 and ending on December 31, 2020, and are useful to form the components of the vectors $I$ and $R$ (hence, of $i$ and $r,$ as in (\ref{fractions})) mentioned in Algorithm \ref{alg2}.

The first case of coronavirus recorded in Brazil occurred on February 26, 2020\footnote{https://agenciabrasil.ebc.com.br/saude/noticia/2025-02/covid-19-primeiro-caso-foi-confirmado-em-sao-paulo-ha-cinco-anos}, with records collected from March onward. Since the numbers of infected and recovered persons during March and April 2020 are too small to exhibit a clear trend without noise, and for data consistency reasons, we decided to perform our analyzes using data collected during the period from May 1, 2020 to December 31, 2020.

In our numerical simulations, we assume that the number of infected persons, $I(t),$ and the number of recovered persons, $R(t),$ are known for days $0\leq t\leq T-1,$ and we calculate the predicted data $\hat{I}(t),$ and $\hat{R}(t),$ according to Algorithm \ref{alg2},  for days $t = T, T+1,...,T+W-1,$ where $T$ is the period that describes the window of known data and $W$ describes, as we said before, the window of predicted data. To know how close the estimated data we obtain are to the real data, we analyze both approximation errors defined by
\begin{equation}
\label{errors}
    \textrm{err}_{W} (I,\hat{I}) =  \frac{\left\| I - \hat{I}\right\|_{\infty}}{\left\| I \right\|_{\infty}},\,\, \, \textrm{err}_{W} (R,\hat{R}) = \frac{\left\| R - \hat{R}\right\|_{\infty}}{\left\| R \right\|_{\infty}}.
\end{equation}
Here, $\| \cdot \|_{\infty}$ is the norm of the maximum when $t \in \{T, T+1,\,  ... ,\,  T+W-1\},$ i.e. 
\begin{equation*}
\left\| I - \hat{I}\right\|_{\infty} = \max\{| I(t)-\hat{I}(t)|, \,\,\,\, T \leq t \leq T+W-1 \}, \,\,\, \textrm{etc.} 
\end{equation*}

We set the values $T=45$ and $W=7.$ If we choose a given month that contains 30 days to perform our numerical simulation, the day $t=0$ corresponds to the first day of this month and the day $t = T-1 = 44$ corresponds to the fifteenth day of the following month. If the chosen month has 31 days, then the day $t=44$ is the fourteenth day of the following month. In our simulation, we assign to $I(t), t = 0, 1, ..., 44,$ the respective data stored in the first column of the table mentioned in the first paragraph of this subsection. In an analogous manner, we use the respective data from the second column to form $R(t), t = 0, 1, ..., 44.$

For the predictions $\hat{\beta}(t),\hat{\gamma}(t), \hat{I}(t+1),$  and $\hat{R}(t+1)$ (follow steps from 4 to 12 in Algorithm \ref{alg2}) for the days $t  \in \{T-1, T, ..., T+W-2 \} = \{ 44, 45, ..., 50\} $ of each of the last seven months of 2020, we take into account in our numerical simulation different values assigned to the orders of the FIR filters $J$ and $K$ and to the regularization parameters $\alpha_1$ and $\alpha_2.$ For $J$ and $K,$ we consider not only the values tested by
Chen  \textit{et al.} \cite{CHEN} ($J = K =3$), but also $J = K = \textrm{round}(T/4) = 11,$ where round$(x)$ is an internal function of the Octave software that returns the integer closest to $x.$ For the pair $(\alpha_1 , \alpha_2),$ we consider all thirty-six combinations of values when
\begin{equation}
\label{parameters}
    \alpha_k \in \{ 10^{-n}, n = 1,2,...,6\}, \,\,\, \textrm{for} \,\,\, k = 1, 2,
\end{equation}
as well as the values tested by Chen  \textit{et al.} \cite{CHEN} which are $\alpha_1 = 0.03$ and $\alpha_2 = 10^{-6}.$

The months of May and June are the only months in which, regardless of the choices discussed above for the regularization parameters $\alpha_1, \alpha_2,$ and for the FIR filters $J, K,$ all the values obtained for the error $\textrm{err}_{W} (I,\hat{I})$ are of order $10^{-1},$ and the values for $\textrm{err}_{W} (R,\hat{R})$ are of order $10^{-2}.$ For the remaining months, the experiments demonstrate that $J = K = 11$ mostly produces better approximations for the estimates of the number of infected and recovered persons than $J = K = 3,$ regardless of the values assigned to the regularization parameters. The experiments also show that when we use the same values tested by Chen \textit{et al.} \cite{CHEN}, i.e., 
\begin{equation}
\label{chenchoices}
  \alpha_1 = 0.03, \,  \alpha_2 = 10^{-6}, \, J = K =3,
\end{equation}
in most of our simulations, the approximations for the estimates $\hat{I}$ and $\hat{R}$ are not as good as those obtained when we use $J = K = 11$ and any of the thirty-six possible choices in (\ref{parameters}).

In the face of good and bad approximations obtained through different choices for the regularization parameters and FIR filters in different months, it becomes necessary to choose a single value for each parameter and filter that serves to produce the good approximations in the simulations performed in most months. In fact the experiments show us that the choices of
\begin{equation}
  \alpha_1 = 10^{-3}, \, \alpha_2 = 10^{-4}, \, J = K = 11 \label{choices}
\end{equation} are the ones that produce good approximation errors, all of them of order $10^{-3},$ for all months from August to November.

Table \ref{tab2} (respectively Table \ref{tab3}) shows the comparison between the approximation errors $\textrm{err}_{W} (I,\hat{I})$ (respectively $\textrm{err}_{W} (R,\hat{R})$), obtained when we run the original algorithm of Chen \textit{et al.} \cite{CHEN} (with their choices for the regularization parameters and FIR filters given as in (\ref{chenchoices})) and our Algorithm \ref{alg2} (with our choices given in (\ref{choices})). 
In these tables, we display the error values within the seven-day prediction window (days $t=45, 46, ..., 51,$) for each simulation performed in which $t=0$ is the first day of a given chosen month. For example, if the chosen month is May 2020, then the days $t=45, 46, ..., 51,$ correspond to the days from June 15 to 21. If we choose a new month for a new simulation, like June, then the days $t=45, 46, ..., 51,$ correspond to the days from July 16 to 22 and so on. As we can see in these tables, in most months, the approximation errors obtained with our algorithm present better values, in the order of $10^{-3},$ compared to the values obtained by Chen  \textit{et al.} \cite{CHEN}, in the order of $10^{-2}.$

\begin{table}
\caption{\textbf{Calculation of approximation error $\textrm{err}_{W} (I,\hat{I}),$ using (\ref{errors}), where $\hat{I}(t)$ is obtained according to Original Method (by \cite{CHEN}), and Modified Method (by Algorithm \ref{alg2}).}}
\setlength{\tabcolsep}{3pt}
\begin{tabular}{|p{45pt}|p{85pt}|p{55pt}|p{60pt}|}
\hline
 Starting on the first day of the month: & 
Forecast days \par ($W=7$): \par $t = 45, 46, \dots , 51$ &
\textsc{Original Method} \par $ J = K = 3, $ \par $\alpha_1 = 0.03,$ \par $ \alpha_2 = 10^{-6}.$ & \textsc{Modified Method} \par $ J = K = 11, $ \par $\alpha_1 = 10^{-3},$ \par $ \alpha_2 = 10^{-4}.$ \\
\hline
May       &  Jun. 15 to Jun. 21  &  0.137886   &  0.110592  \\ \hline
June      &  Jul. 16 to Jul. 22  &  0.166842   &  0.176757  \\ \hline
July      &  Aug. 15 to Aug. 21  &  0.076807   &  0.018101  \\ \hline    
August    &  Sep. 15 to Sep. 21  & 0.045647    &   6.7120e-03  \\ \hline
September &  Oct. 16 to Oct. 22  & 0.018525    &   2.9539e-03  \\ \hline
October   &  Nov. 15 to Nov. 21  & 5.0550e-03  &   3.7620e-03  \\ \hline
November  &   Dec. 16 to Dec. 22 & 0.019479    &   6.3362e-03  \\ \hline
\end{tabular}
\label{tab2}
\end{table}

\begin{figure}[hbtp]
    \centering
   \includegraphics[width=1.0\linewidth]{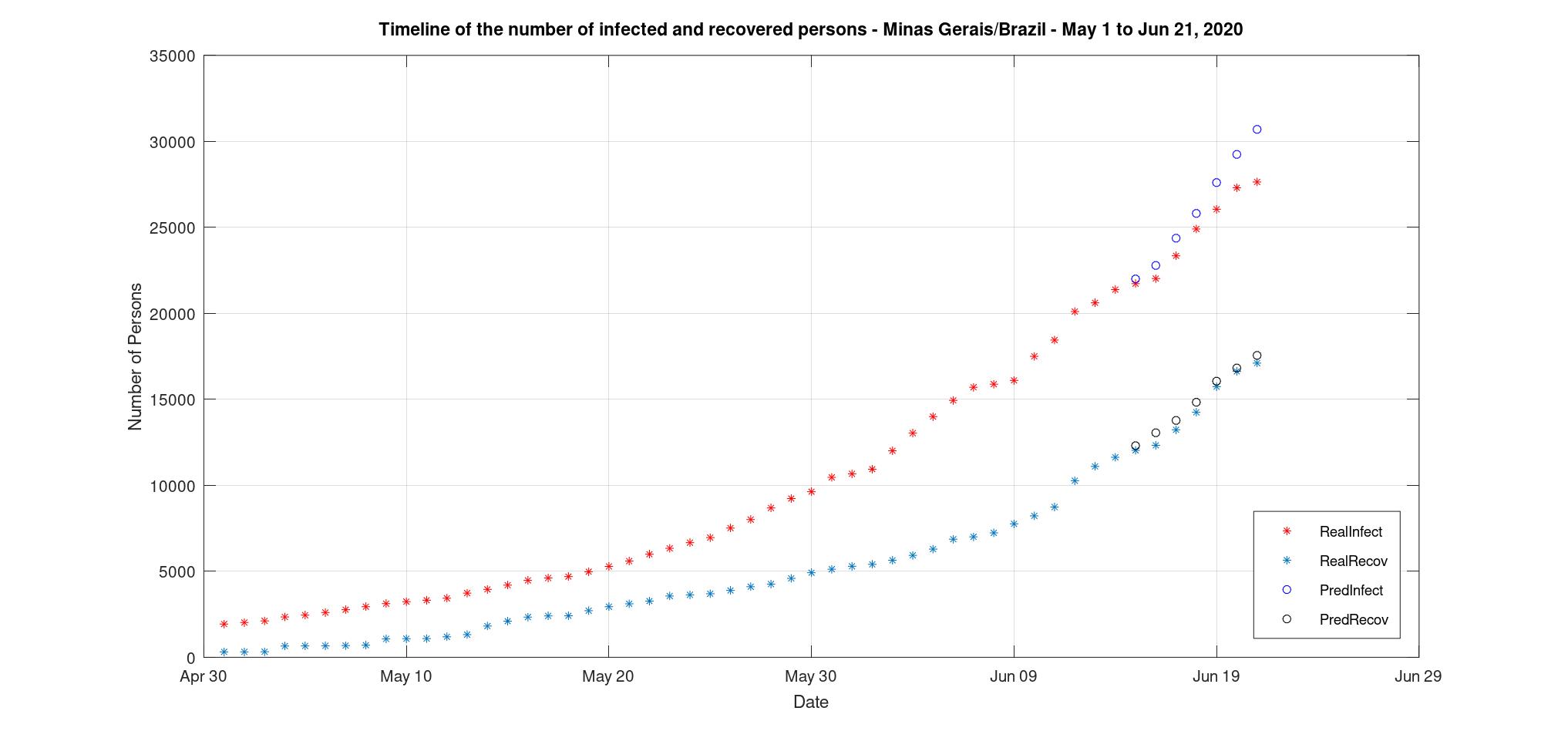} \\
   \includegraphics[width=1.0\linewidth]{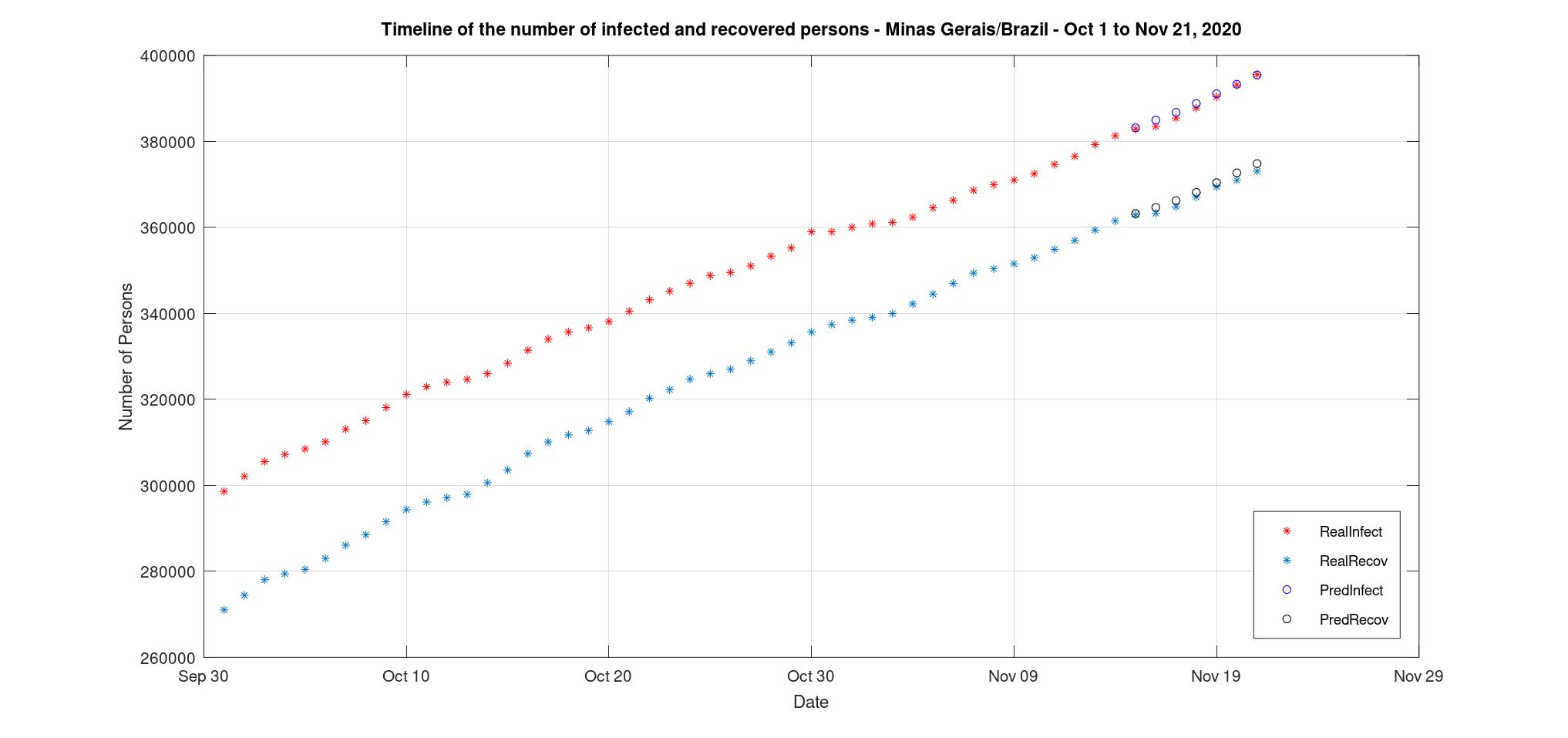}
    \caption{\tiny Timeline of the time-dependent SIR model of COVID-19 for the state of Minas Gerais, Brazil. The first (resp. second) graph shows the evolution from May 1, 2020 to Jun. 21, 2020 (resp. from Oct. 1, 2020 to Nov. 21, 2020). The curve marked with red (resp. cyan) stars is the real number of infected persons $I(t)$ (resp. recovered persons $R(t)$). The curve marked with blue (resp. black) circles is the predicted number of infected persons $\hat{I}(t)$  (resp. recovered persons $\hat{R}(t)$) within the seven day prediction window, days from Jun. 15 to Jun. 21 (resp. from Nov. 15 to Nov. 21). We compute the predictions $\hat{I}(t)$ and $\hat{R}(t)$ by using Algorithm \ref{alg2}, with FIR filters $J = K = 11$ and regularization parameters $\alpha_1 = 10^{-3}, \alpha_2 = 10^{-4}$. The approximation errors are $\textrm{err}_{W} (I,\hat{I}) = 0.110592$ and $\textrm{err}_{W} (R,\hat{R}) = 0.042655$ (resp. $\textrm{err}_{W} (I,\hat{I}) = 3.7620\textrm{e-03}$ and $\textrm{err}_{W} (R,\hat{R}) = 4.6717\textrm{e-03}$). }
    \label{may_oct}
\end{figure}

\begin{figure}[hbtp]
    \centering
   \includegraphics[width=1.0\linewidth]{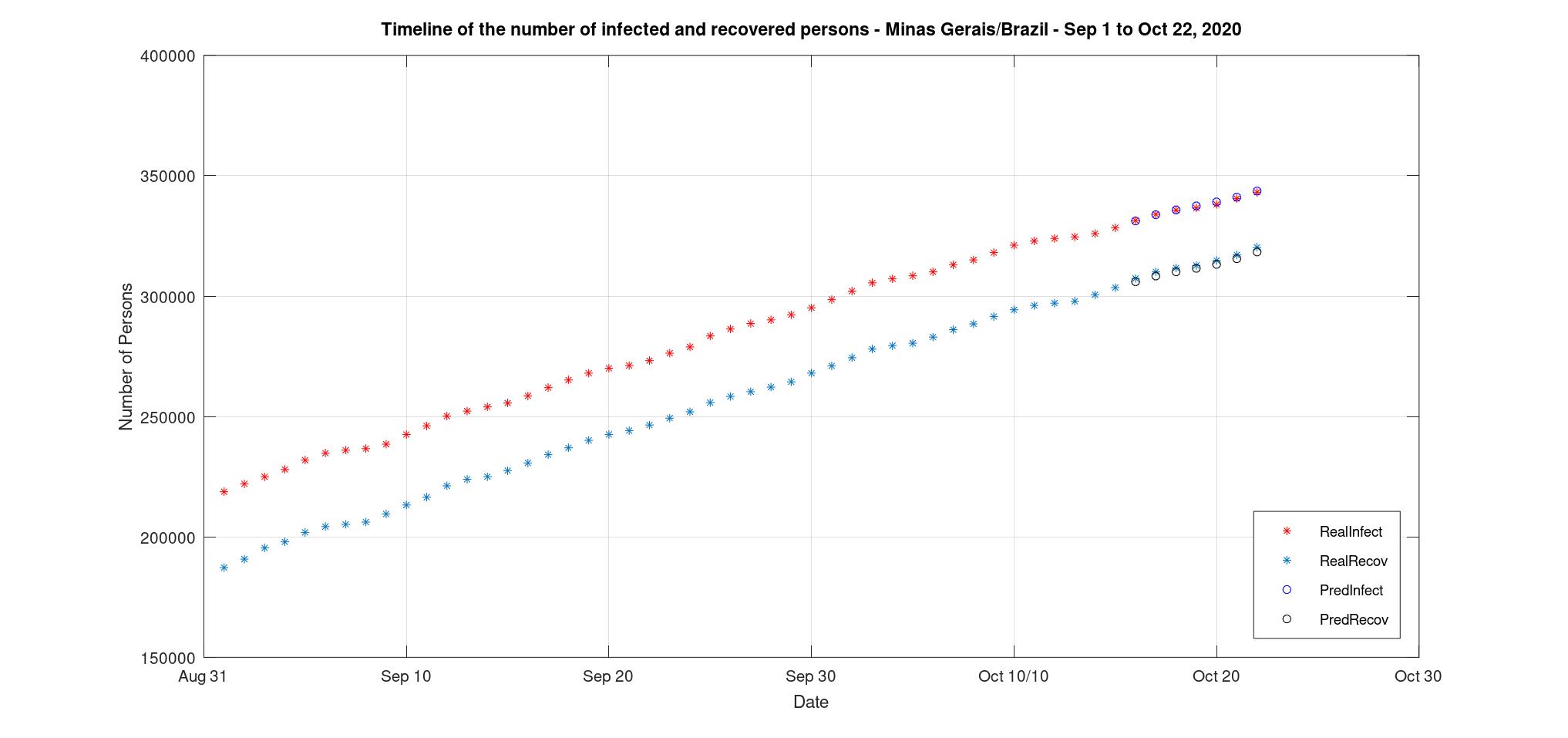} \\
   \includegraphics[width=1.0\linewidth]{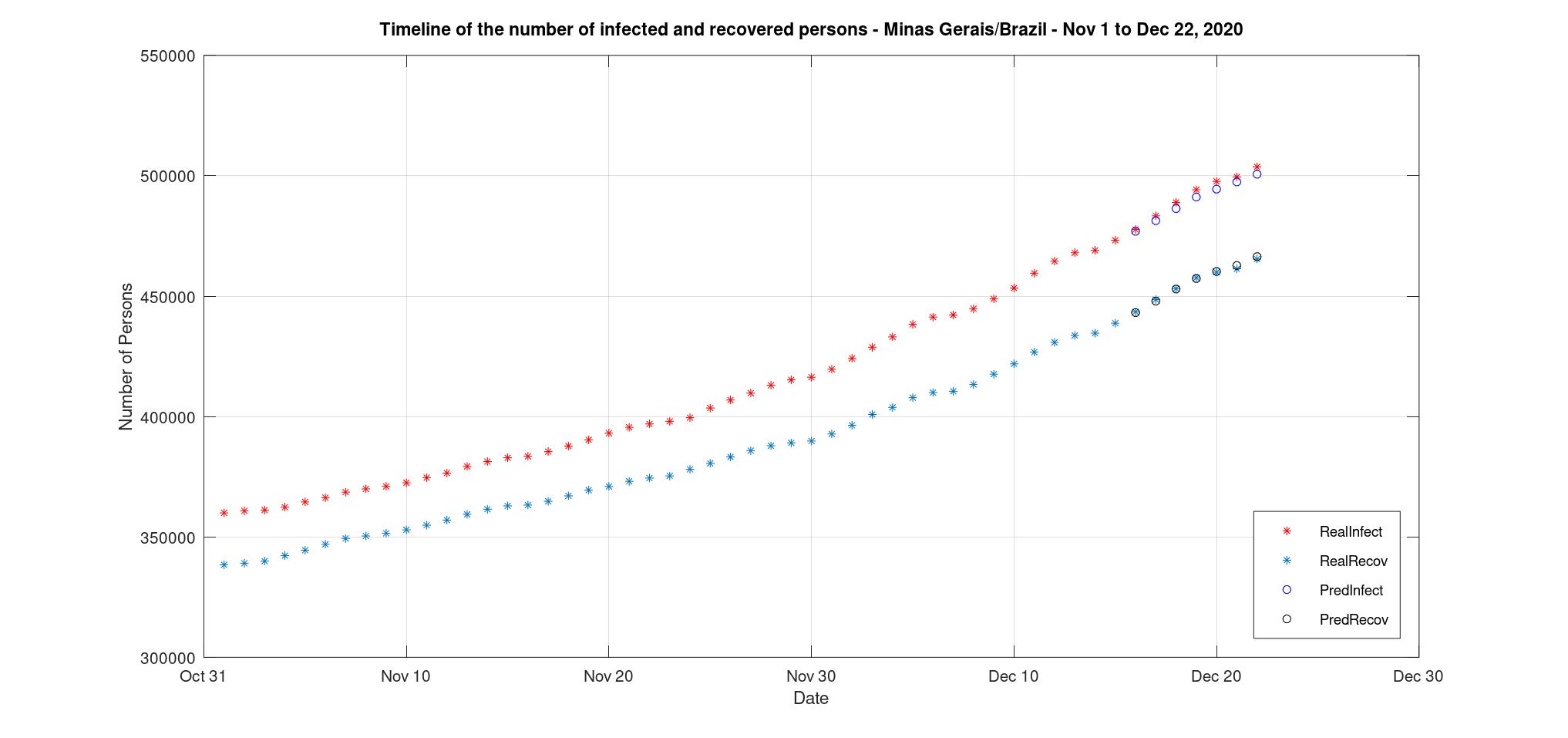} 
    \caption{\tiny Timeline of the time-dependent SIR model of COVID-19 for the state of Minas Gerais, Brazil. The first (resp. second) graph shows the evolution from Sep. 1, 2020 to Oct. 22, 2020 (resp. from Nov. 1, 2020 to Dec. 22, 2020). The curve marked with red (resp. cyan) stars is the real number of infected persons $I(t)$ (resp. recovered persons $R(t)$). The curve marked with blue (resp. black) circles is the predicted number of infected persons $\hat{I}(t)$  (resp. recovered persons $\hat{R}(t)$) within the seven day prediction window, days from Oct. 16 to Oct. 22 (resp. from Dec. 16 to Dec. 22). We compute the predictions $\hat{I}(t)$ and $\hat{R}(t)$ by using Algorithm \ref{alg2}, with FIR filters $J = K = 11$ and regularization parameters $\alpha_1 = 10^{-3}, \alpha_2 = 10^{-4}$. The approximation errors are $\textrm{err}_{W} (I,\hat{I}) = 2.9539\textrm{e-03}$ and $\textrm{err}_{W} (R,\hat{R}) = 5.9718\textrm{e-03}$ (resp. $\textrm{err}_{W} (I,\hat{I}) = 6.3362\textrm{e-03}$ and $\textrm{err}_{W} (R,\hat{R}) = 3.0063\textrm{e-03}$). }
    \label{sep_nov}
\end{figure}

\begin{table}
\caption{\textbf{Calculation of approximation error $\textrm{err}_{W} (R,\hat{R}),$ using (\ref{errors}), where $\hat{R}(t)$ is obtained according to Original Method (by \cite{CHEN}), and Modified Method (by Algorithm \ref{alg2}).}}
\setlength{\tabcolsep}{3pt}
\begin{tabular}{|p{45pt}|p{85pt}|p{55pt}|p{60pt}|}
\hline
 Starting on the first day of the month: & 
Forecast days \par ($W=7$): \par $t = 45, 46, \dots , 51$ &
\textsc{Original Method} \par $ J = K = 3, $ \par $\alpha_1 = 0.03,$ \par $ \alpha_2 = 10^{-6}.$ & \textsc{Modified Method} \par $ J = K = 11, $ \par $\alpha_1 = 10^{-3},$ \par $ \alpha_2 = 10^{-4}.$ \\
\hline
May       &  Jun. 15 to Jun. 21  &  0.031318   &   0.042655  \\ \hline
June      &  Jul. 16 to Jul. 22  &  0.082418   &   0.093887  \\ \hline
July      &  Aug. 15 to Aug. 21  &  0.074709   &   0.017556  \\ \hline    
August    &  Sep. 15 to Sep. 21  & 0.032734    &   8.6285e-03  \\ \hline
September &  Oct. 16 to Oct. 22  & 0.017313    &   5.9718e-03 \\ \hline
October   &  Nov. 15 to Nov. 21  & 0.012043    &   4.6717e-03  \\ \hline
November  &   Dec. 16 to Dec. 22 & 0.014563    &   3.0063e-03 \\ \hline
\end{tabular}
\label{tab3}
\end{table}
Figures \ref{may_oct} and \ref{sep_nov} show four simulations, all of which are obtained by executing Algorithm \ref{alg2}. They describe the evolution of the number of people infected and recovered in 52 days. Each simulation considers the initial day of evolution (day $t=0$) as the first day of the respective months May, October, September, and November. There is no special reason for choosing these months, other than that we want to show months that present approximation errors of orders $10^{-1}, 10^{-2},$ and $10^{-3}$. For all four of these simulations, we assume the filter and parameter values as in (\ref{choices}).
The curve marked with red (resp. cyan) stars is the real number of infected persons $I(t)$ (resp. recovered persons $R(t)$) over 52 days. The curve marked with blue (resp. black) circles is the predicted number of infected persons $\hat{I}(t)$ (resp. recovered persons $\hat{R}(t)$) during the 7-day prediction window (days $t=45, 46, \dots , 51$). As we can see in Figs. \ref{may_oct} and \ref{sep_nov}, the simulation for the month of May shows that the predictions $\hat{I}(t)$ and $\hat{R}(t)$ are a little further from the real values $I(t)$ and $R(t),$ within the prediction window, compared to the simulations for the other months. This is reflected in the accuracy of the approximation errors. In fact, while the approximation errors $\textrm{err}_{W} (I,\hat{I})$ and $\textrm{err}_{W} (R,\hat{R})$ for the simulation in May present values of order $10^{-1}$ and $10^{-2},$ respectively, the error values for the simulations in other months are all of order $10^{-3}$. 

\begin{figure}[hbtp] 
    \centering
   \includegraphics[width=1.0\linewidth]{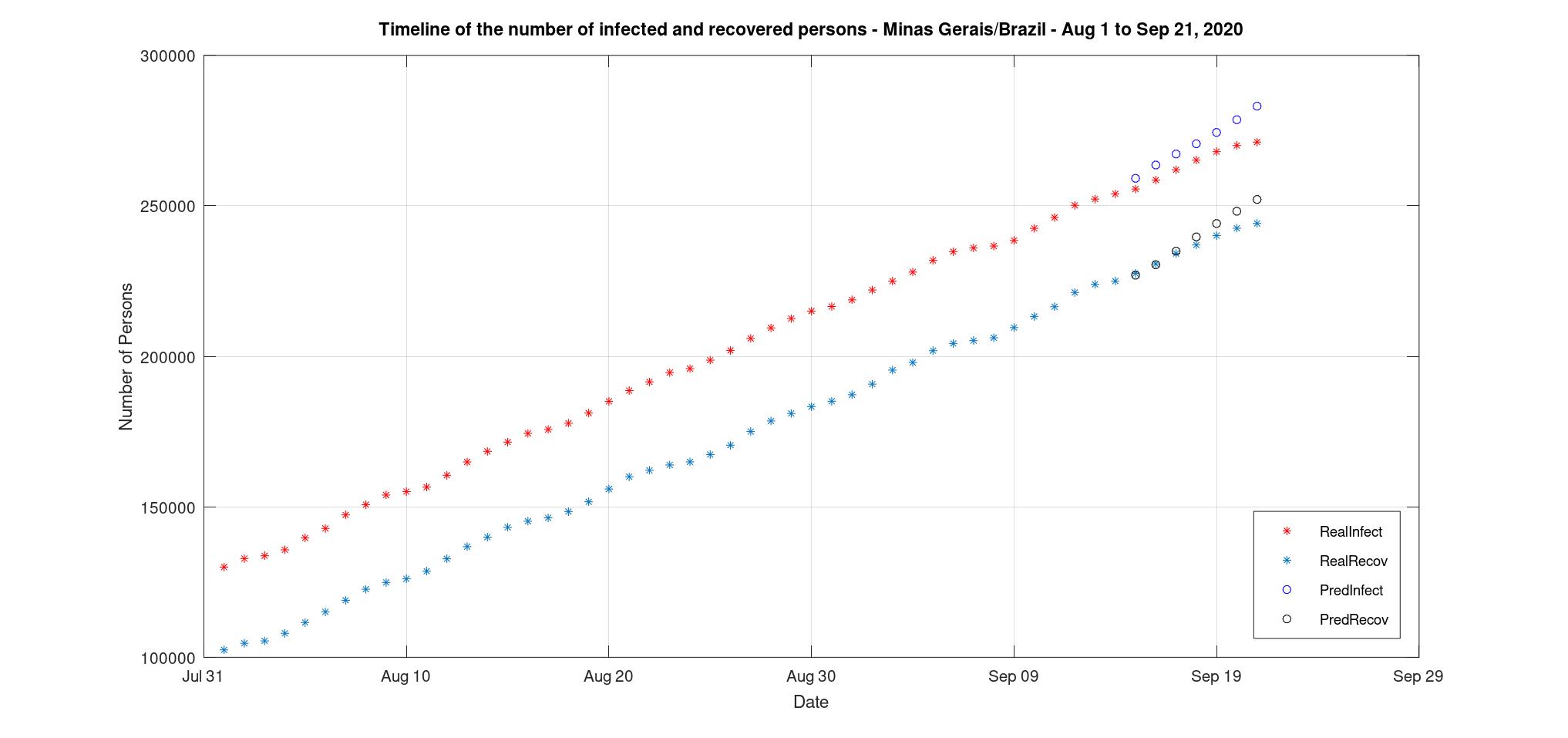} \\
   \includegraphics[width=1.0\linewidth]{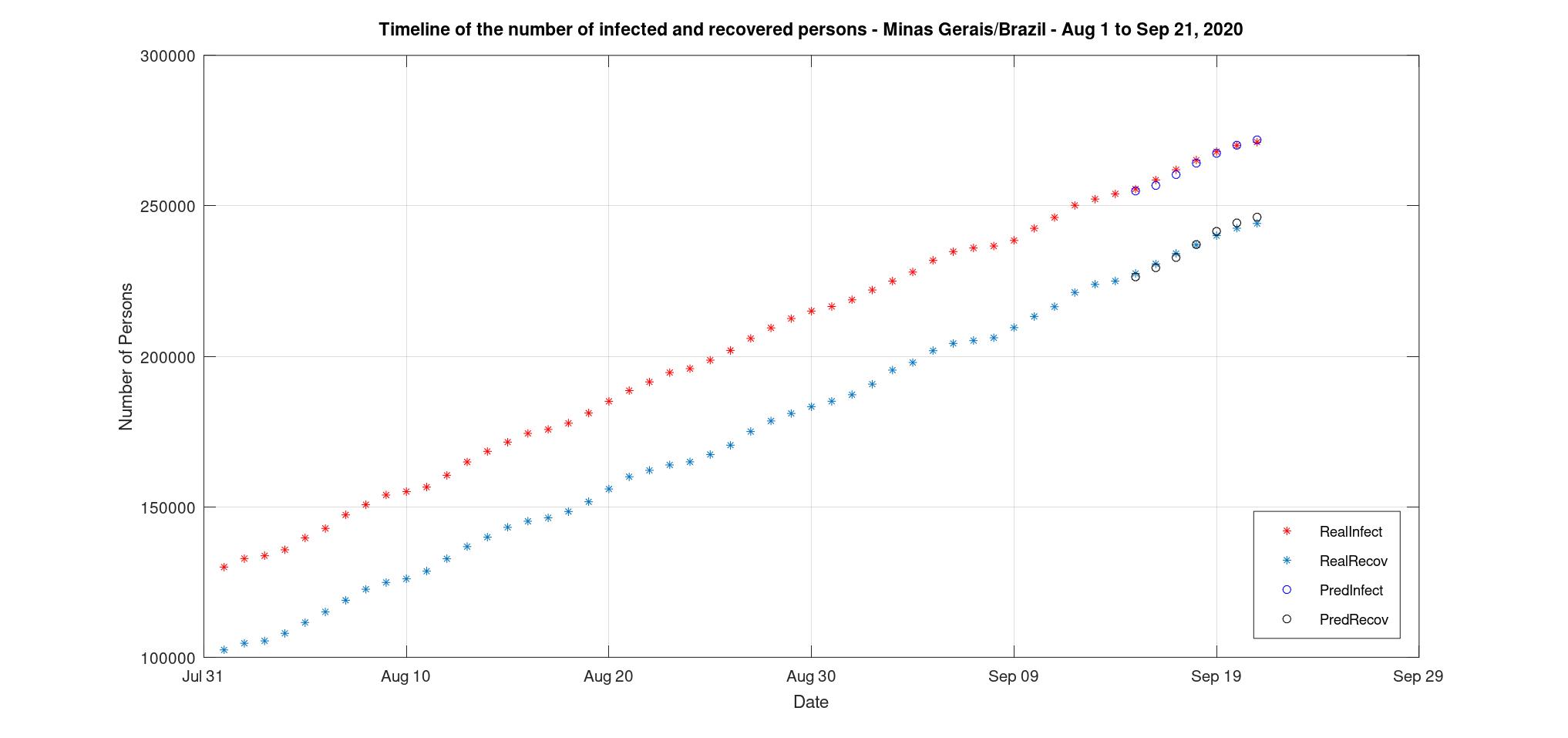} 
    \caption{\tiny Timeline of the time-dependent SIR model of COVID-19 for the state of Minas Gerais, Brazil. Both graphs show the evolution from Aug. 1, 2020 to Sep. 21, 2020. The curve marked with red (resp. cyan) stars is the real number of infected persons $I(t)$ (resp. recovered persons $R(t)$). The curve marked with blue (resp. black) circles is the predicted number of infected persons $\hat{I}(t)$  (resp. recovered persons $\hat{R}(t)$) within the seven day prediction window, days from Sep. 15 to Sep. 21. In the graph above, the predictions of the numbers of infected and recovered people were obtained by executing the algorithm of Chen  \textit{et al.} \cite{CHEN} with regularization parameters $\alpha_1 = 0.03, \, \alpha_2 = 10^{-6}$ and FIR filters $J = K = 3.$ In this case, the approximation errors are $\textrm{err}_{W} (I,\hat{I}) = 0.045647$ and $\textrm{err}_{W} (R,\hat{R}) = 0.032734.$ In the graph below, the respective predictions were obtained by executing Algorithm \ref{alg2} with regularization parameters $\alpha_1 = 10^{-3},$ $\alpha_2 = 10^{-4},$ and FIR filters $J = K = 11.$ In this case, the approximation errors improve, with values $\textrm{err}_{W} (I,\hat{I}) = 6.7120\textrm{e-03}$ and $\textrm{err}_{W} (R,\hat{R}) = 8.6285\textrm{e-03}.$ }
    \label{aug}
\end{figure}

We conclude this subsection by comparing the simulations obtained for the month of August when we run the original method and the modified method. As we can see from the error values presented in Tables 2 and 3, the original method of Chen \textit{et al.} \cite{CHEN} with the parameters and filters given as in (\ref{chenchoices}) does not produce approximations for the estimates $\hat{I}(t)$ and $\hat{R}(t)$ as good as those obtained when we use our modified method with the parameters and filters given as in (\ref{choices}). This observation can be seen in Fig. 3 where we show the evolution of the number of infected and recovered persons over fifty-two days, counting from August 1st. The upper (resp. lower) graph in Fig. \ref{aug} shows the real and estimated data when we use the original method by Chen \textit{et al.} \cite{CHEN} (resp. Algorithm \ref{alg2}) with parameters $\alpha_1 = 0.03,\, \alpha_2 = 10^{-6}$ (resp. $\alpha_1 = 10^{-3}, \, \alpha_2 = 10^{-4}$), and filters $J = K = 3$ (resp. $J = K = 11$). When we run the algorithm of Chen \textit{et al.} \cite{CHEN}, the approximation errors are $\textrm{err}_{W} (I,\hat{I}) = 0.045647$ and $\textrm{err}_{W} (R,\hat{R}) = 0.032734,$ and when we run the Algorithm \ref{alg2}, the approximation errors improve, with values $\textrm{err}_{W} (I,\hat{I}) = 6.7120\textrm{e-03}$ and $\textrm{err}_{W} (R,\hat{R}) = 8.6285\textrm{e-03}.$
\section{Concluding Remarks}
\label{conclusion}

As we saw in Subsection (\ref{numericalresults}), the modified method presented in this article produced, in most of the months tested between May and December 2020, better approximation errors compared to the errors presented by the method of Chen \textit{et al.} \cite{CHEN}. However, the comparison between these two methods could not be continued throughout the years 2021 and 2022 because these were the initial years in which the COVID-19 vaccination campaign took place in Brazil, and the time-dependent discrete SIR model described by equations (\ref{discrete_s}), (\ref{discrete_i}), and (\ref{discrete_r}) does not take vaccination into account. To include vaccination in our model, we would have to add a new compartment of vaccinated individuals and at least one more new parameter, in addition to the parameters $\beta(t)$ and $\gamma(t).$ If we take into account a single additional parameter, say $\rho (t),$ that describes the vaccination rate of susceptible individuals, we generate a new discrete SIRV-type model, but applicable in a vaccination context in which only people who were not infected by the disease took the vaccine. In this simplified SIRV model we can also isolate $\rho (t)$ in terms of the other variables of the model, as we observe with the parameters $\beta(t)$ and $\gamma(t)$ in equations (\ref{gammat}) and (\ref{betat}), so that $\beta, \gamma$ and $\rho$ are independent of each other.
\begin{quote}
    \textit{Note}: Equations (\ref{gammat}) and (\ref{betat}) show us that the parameters $\beta(t)$ and $\gamma(t)$ are independent of each other in the sense that both expressions on the right-hand side of (\ref{gammat}) and (\ref{betat}) do not involve any model parameter, but only the variables that describe the number of individuals in the model compartments over the total population.
\end{quote}
 However, if we take into account that vaccinated individuals do not have permanent immunity, then a new parameter that describes the average contact rate of an infectious individual with a vaccinated individual will also have to be added to the model. Furthermore, we would have to separate the compartment of those who died from the disease from the compartment of those who recovered, since people who died from the disease are not vaccinated. This makes the model increasingly complex, and the technique of isolating parameters that are independent of each other becomes a difficult, if not impossible, task. 

We are currently working on a model that predicts vaccination for both susceptible individuals who have never been infected and individuals who have had the disease, become temporarily immune, and then return to the susceptible class. Our model should assume not only a single-dose immunization scenario but also a multiple-dose immunization scenario. To overcome the difficulties that arise from the interdependence of the parameters involved, one possibility is to try to obtain some of these parameters directly from the Brazilian Ministry of Health database, if they are available.

\par
\section*{About the authors}
\par
\textbf{Felipe Rog\'erio Pimentel} is currently a full professor in the Department of Mathematics of the Federal University of Ouro Preto, Ouro Preto, Minas Gerais, Brazil.
\newline\newline
\textbf{Rafael Gustavo Alves} is a former student of the Federal University of Ouro Preto, where he obtained a bachelor's degree in mathematics. Part of the content of this article is based on his final project, supervised by Professor Felipe R. Pimentel. He currently works as a teacher at the Izaura Mendes Elementary School in Ouro Preto, Minas Gerais, Brazil.

\end{document}